\PassOptionsToPackage{numbers,compress}{natbib}
\documentclass{article}

\usepackage[main, final]{neurips_2026}
\usepackage[utf8]{inputenc} % allow utf-8 input
\usepackage[T1]{fontenc}    % use 8-bit T1 fonts
\usepackage{url}            % simple URL typesetting
\usepackage{booktabs}       % professional-quality tables
\usepackage{amsfonts}       % blackboard math symbols
\usepackage{nicefrac}       % compact symbols for 1/2, etc.
\usepackage{microtype}      % microtypography
\usepackage{xcolor}         % colors
\usepackage{graphicx}       % resizebox
\usepackage{amsmath}
\usepackage{algorithm}
\usepackage{algorithmic}
\usepackage{enumitem}
\usepackage{hyperref}       % hyperlinks; load last (after float/algorithm) to avoid duplicate PDF anchors
\title{End-to-End Differentiable Diffusion Conditioning for Physics-Informed Optimization}

\author{%
  Ricardo Luna Gutierrez$^1$$^\dag$, Vineet Gundecha$^1$$^\dag$, Rahman Ejaz$^2$$^\dag$, Varchas Gopalaswamy$^2$$^\dag$, \\ 
  \textbf{Riccardo Betti}$^2$$^\dag$, \textbf{Sahand Ghorbanpour}$^1$$^\dag$, \textbf{Aarne Lees}$^2$$^\dag$, \textbf{Soumyendu Sarkar}$^{1}$$^\dag$\thanks{Corresponding author. \dag These authors contributed equally.} \\
  $^1$Hewlett Packard Enterprise, $^2$University of Rochester \\
  \texttt{\{rluna, vineet.gundecha, sahand.ghorbanpour, soumyendu.sarkar\}@hpe.com} \\
  \texttt{\{reja, vgop, betti, alees\}@lle.rochester.edu}
}

\begin{document}

\maketitle

\begin{abstract}
Optimizing high-dimensional physical systems under expensive evaluations and
complex constraints is a central challenge in science and engineering. In
offline model-based optimization (MBO), methods must improve designs using only
a fixed dataset and no test-time access to the true objective. However,
existing methods typically suffer from conservatism, unconstrained surrogate
exploitation, or function as static inference-time samplers that cannot refine
candidates against precise physical objectives. In this paper, we propose Differentiable Diffusion 
Conditioning (D2C) for offline MBO, which leverages the fact that deterministic sampling makes the reverse denoising
chain differentiable with respect to the conditioning signal. D2C turns a
frozen conditional diffusion model into an end-to-end gradient-based optimizer
by directly optimizing conditioning variables at test time via gradients from
frozen surrogate objectives and physics-informed penalties backpropagated
through the reverse chain. We implement this idea in
a composed proposer-evaluator framework and evaluate it on laser pulse
optimization for inertial confinement fusion, sustainable data-center workload
scheduling, and four Design-Bench tasks. Our method achieves the strongest
results on the two real-world tasks and the best average rank across four
Design-Bench tasks.
\end{abstract}

\section{Introduction}

The optimization of complex physical systems remains one of the most consequential challenges in science and engineering.
Problems such as laser pulse shaping for inertial confinement fusion (ICF) and workload scheduling in large-scale data centers share a common structure: objective functions are computationally expensive to evaluate, the design space is high-dimensional with non-linear correlations, and candidate solutions must satisfy strict physical constraints where any violation carries safety-critical consequences.

The offline model-based optimization (MBO) paradigm addresses the evaluation-cost bottleneck by learning entirely from a fixed dataset of previously evaluated designs, with no access to the true objective at test time.
Within this paradigm, existing approaches fall into three broad families.
Forward (surrogate-based) methods train a neural network to approximate the objective and perform gradient ascent in design space~\citep{trabucco2021coms, fu2021nemo}; while effective, optimizing directly in unconstrained design space can move into regions poorly supported by the offline data, where the surrogate is unreliable.
Inverse methods learn a conditional mapping from target performance labels to designs~\citep{brookes2019cbas, kumar2020mins, krishnamoorthy2023ddom, gutierrez2025imlpg}; these tend to stay closer to the empirical data distribution but are limited by the performance ceiling of the training labels and struggle to extrapolate beyond observed optima.
Trajectory-based methods construct synthetic improvement paths and train generative models to imitate optimization dynamics~\citep{chemingui2024pgs, yun2024gtg}; while powerful, they remain inference-time samplers whose conditioning targets are fixed prior to generation, with no mechanism to refine designs against downstream objectives.

A common limitation across these families is the lack of a direct, differentiable path from a physics-informed objective back to the generative process.
The conditioning signal that steers generation is treated as a static input rather than an optimizable variable.
This decoupling is especially limiting in safety-critical physical domains where the objective is not a single scalar score but a composite of performance metrics, conservation laws, and hard feasibility constraints that must be jointly satisfied.

We address these limitations by proposing Differentiable Diffusion Conditioning (D2C) for Physics-Informed Optimization, a novel approach that converts a conditional diffusion model into an end-to-end gradient-based optimizer for constrained physical design.
Our key insight for offline MBO is that deterministic diffusion sampling renders the reverse chain differentiable with respect to the conditioning input, enabling end-to-end backpropagation from downstream physics-informed objectives, through a frozen surrogate evaluator and physics-informed penalty terms, back to the conditioning variables.
D2C therefore optimizes conditioning directly at test time while keeping both the diffusion proposer and surrogate evaluator fixed, yielding a practical end-to-end gradient path from the downstream objective to the conditioning variables. The diffusion model tends to keep candidates closer to the data distribution, while surrogate and physics-informed losses steer them toward feasible, high-performing designs.

\begin{figure}[tbp] % [htbp] for more placement flexibility
    \vskip -0.4cm
    \centering
    \includegraphics[width=0.9\textwidth]{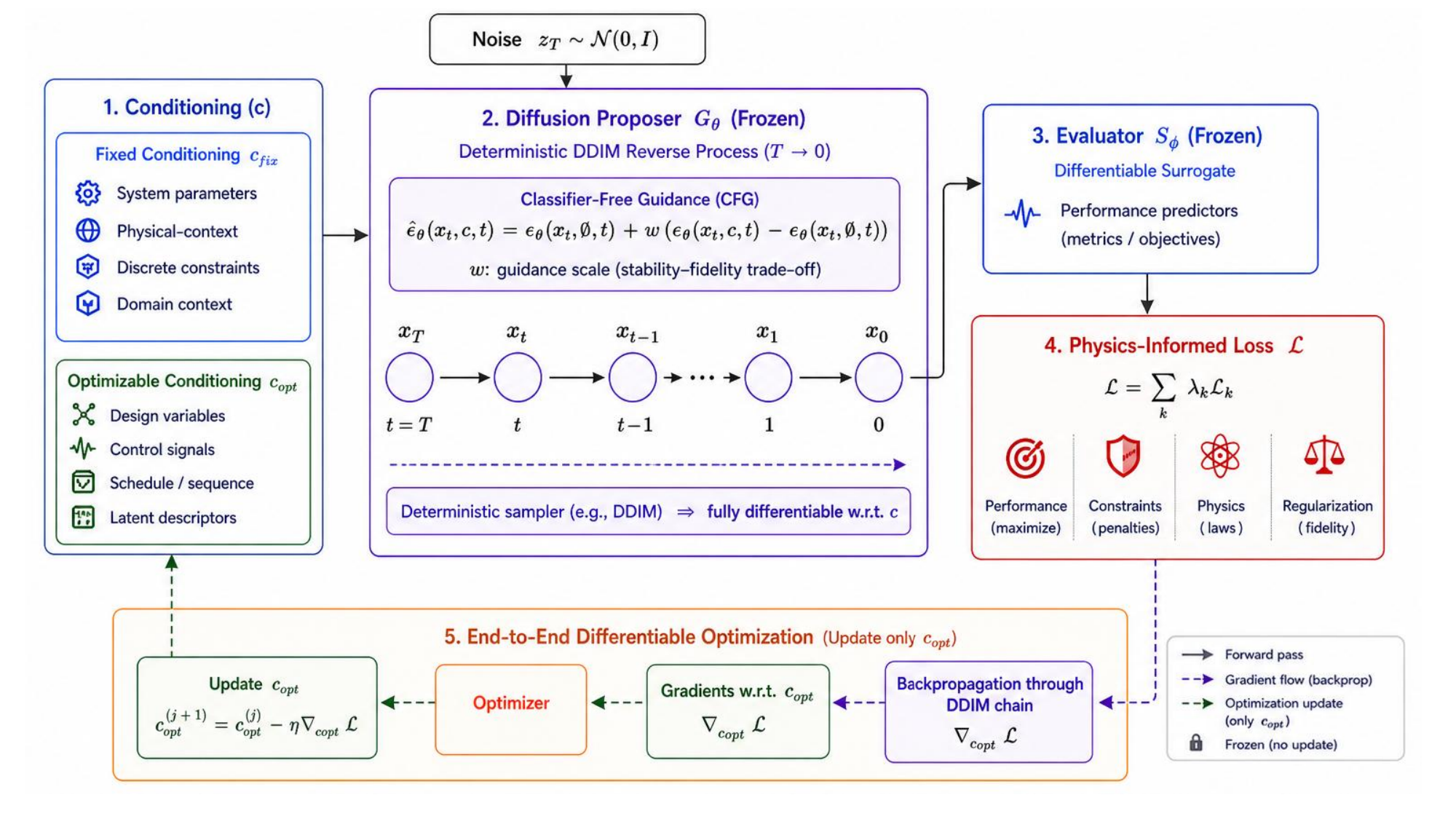}
       
    \caption{Overview of D2C. A frozen conditional diffusion proposer generates designs from conditioning inputs, a frozen surrogate evaluator and physics-informed penalties define a differentiable test-time objective, and gradients are backpropagated through the deterministic DDIM chain to optimize the conditioning signal. Multiple noise trials and loss-weight samples produce a candidate pool, from which the best valid design is selected after domain-specific constraint checks.}
    \label{fig:overview}
    \vskip -0.2cm
\end{figure}

We embed this mechanism within a composed proposer-evaluator architecture: the diffusion model $G_\theta$ (proposer) generates candidates, and a surrogate $S_\phi$ (evaluator) scores them, with both remaining fixed during test-time optimization.
Only the conditioning input is updated, which helps keep optimization closer to the proposer’s learned data distribution while limiting direct surrogate exploitation.
A stochastic multi-objective weighting scheme and validity-first selection further encourage broad exploration of competing objective trade-offs while enforcing strict constraint satisfaction. Figure~\ref{fig:overview} provides an overview of the full D2C pipeline. Our main contributions are:

\begin{itemize}
    \item \textbf{End-to-End Conditioning Optimization Through Diffusion.} We use the deterministic reverse diffusion chain to backpropagate end-to-end from composite physics objectives to the conditioning inputs, making the conditioning signal itself the optimization variable within the generative process; optimizing it outperforms optimizing the latent noise (Appendix~\ref{sec:ablation_cond_vs_latent}).
    \item \textbf{Frozen Proposer-Evaluator Architecture.} We introduce a composed proposer-evaluator architecture in which a frozen diffusion proposer helps keep candidates closer to the empirical data distribution, while a frozen surrogate evaluator provides optimization gradients, reducing out-of-distribution exploitation.
    \item \textbf{Physics-Informed Multi-Objective Optimization.} We integrate heterogeneous objectives directly into the differentiable sampling chain via a stochastic Dirichlet-weighted loss, enabling principled multi-objective optimization under domain-specific constraints.
    \item \textbf{Strong Empirical Performance.} D2C achieves the best results on the two real-world physical design tasks and the best average rank across four Design-Bench tasks.
\end{itemize}

\section{Preliminaries}

\subsection{Conditional Denoising Diffusion Models}
Denoising Diffusion Probabilistic Models (DDPMs) \citep{ho2020denoising} learn a generative model of a data distribution $p(x)$ by reversing a forward noise process. In the conditional setting, the model learns to sample $x \sim p(x|c)$ given a conditioning signal $c \in \mathbb{R}^{d_c}$. For a continuous design space $x \in \mathbb{R}^d$, the forward process adds Gaussian noise according to a variance schedule $\beta_1, \dots, \beta_T$:
\begin{equation}
q(x_t | x_{t-1}) = \mathcal{N}(x_t; \sqrt{1-\beta_t}x_{t-1}, \beta_t \mathbf{I})
\end{equation}
A noise-prediction network $\epsilon_\theta(x_t, c, t)$ is trained by minimizing the denoising objective:
\begin{equation}
\mathcal{L}_{diff} = \mathbb{E}_{x \sim p(x\mid c),\, \epsilon \sim \mathcal{N}(0,\mathbf{I}),\, t \sim \mathrm{Unif}(\{1,\ldots,T\})} \left[ \| \epsilon - \epsilon_\theta(x_t, c, t) \|^2 \right]
\end{equation}
where $x_t = \sqrt{\bar{\alpha}_t}x + \sqrt{1-\bar{\alpha}_t}\epsilon$ and $\bar{\alpha}_t = \prod_{i=1}^t (1-\beta_i)$. %In this work, $\epsilon_\theta$ is implemented as a Multi-Layer Perceptron (MLP) to efficiently handle high-dimensional continuous design vectors.

\subsection{Deterministic DDIM Sampling}
Denoising Diffusion Implicit Models (DDIM) \citep{song2020denoising} generalize the sampling process by considering a non-Markovian forward process that yields the same marginals as DDPM. Crucially, DDIM allows for a deterministic mapping from a latent noise $x_T$ to a clean sample $x_0$. The update rule is given by:
\begin{equation}\label{eq:ddim_prelim}
x_{t-1} = \sqrt{\bar\alpha_{t-1}} \left( \frac{x_t - \sqrt{1-\bar\alpha_t} \epsilon_\theta(x_t, c, t)}{\sqrt{\bar\alpha_t}} \right) + \sqrt{1-\bar\alpha_{t-1}} \epsilon_\theta(x_t, c, t)
\end{equation}
Eq.~\eqref{eq:ddim_prelim} gives the fully deterministic DDIM update ($\sigma_t = 0$), in which the stochastic noise term is omitted and the sampling chain becomes a deterministic discrete mapping from $x_T$ to $x_0$. This update can be viewed as a discrete analogue of the probability-flow ODE perspective, but here we rely only on the finite-step deterministic map induced by the sampler. This determinism ensures that the mapping $G_\theta: (c, x_T) \mapsto x_0$ is differentiable with respect to the conditioning input $c$.

\subsection{Classifier-Free Guidance (CFG)}
Classifier-Free Guidance is a technique to improve the adherence of the generated samples to the conditioning signal $c$. During training, the conditioning $c$ is randomly replaced with a null token $\emptyset$ with probability $p_{\mathrm{drop}}$. At inference time, the noise estimate is interpolated between the conditional and unconditional predictions:
\begin{equation}
\tilde\epsilon_\theta(x_t, c, t) = \epsilon_\theta(x_t, \emptyset, t) + w (\epsilon_\theta(x_t, c, t) - \epsilon_\theta(x_t, \emptyset, t))
\end{equation}
where $w \geq 0$ is the guidance scale ($w = 0$ yields unconditional sampling, $w = 1$ recovers standard conditional sampling, and $w > 1$ amplifies conditioning adherence). In D2C, we use this guidance not only to bias sampling but also to provide a stable optimization regime in which gradient-based updates of $c$ can proceed.

\section{Methodology}

We present Differentiable Diffusion Conditioning (D2C), a framework that enables a pretrained conditional diffusion model to function as an end-to-end gradient-based optimizer for physical design.
The key idea is a composed proposer-evaluator architecture in which a conditional diffusion model serves as a differentiable proposer and a frozen forward surrogate provides a training-time consistency signal and inference-time gradient guidance.
Algorithm~\ref{alg:d2c} summarizes the full pipeline. We describe each component below.

\subsection{Problem Formulation}

Let $x \in \mathcal{X} \subseteq \mathbb{R}^d$ denote a physical design (e.g.\ a temporal power profile or a laser pulse shape) and $c \in \mathcal{C} \subseteq \mathbb{R}^{d_c}$ its associated conditioning vector encoding environmental state and performance targets.
We decompose the conditioning into a fixed component $c_{\mathrm{fix}}$ (immutable physical context such as ambient conditions) and an optimizable component $c_{\mathrm{opt}}$ (optimizable performance targets):
\begin{equation}
    c = (c_{\mathrm{fix}},\; c_{\mathrm{opt}}), \qquad c_{\mathrm{fix}} \in \mathbb{R}^{d_f},\;\; c_{\mathrm{opt}} \in \mathbb{R}^{d_o},\;\; d_f + d_o = d_c.
\end{equation}
Given a pretrained conditional diffusion model with frozen parameters $\theta$ and a deterministic reverse chain $G_\theta: \mathcal{C} \times \mathbb{R}^d \to \mathcal{X}$ mapping conditioning and a latent noise vector to a design $x_0 = G_\theta(c, x_T)$, we seek, for a \emph{fixed} noise realization $x_T$:
\begin{equation}\label{eq:main_opt}
    c_{\mathrm{opt}}^*(x_T) = \arg\min_{c_{\mathrm{opt}}} \; \mathcal{L}\!\bigl(G_\theta\!\left((c_{\mathrm{fix}},\, c_{\mathrm{opt}}),\, x_T\right),\; x_{\mathrm{ref}},\; c_{\mathrm{fix}}\bigr), \qquad x_T \sim \mathcal{N}(0, \mathbf{I}).
\end{equation}
Each draw $x_T$ induces a different deterministic mapping $G_\theta(\cdot, x_T)$; the optimization is therefore conditional on the noise realization.
To cover multiple modes, we solve Eq.~\eqref{eq:main_opt} independently for $N_z$ draws and select the best feasible solution (Section~\ref{sec:coverage}).
Here $x_{\mathrm{ref}}$ is the reference (baseline) design and $\mathcal{L}$ is a composite physics-informed loss detailed in Section~\ref{sec:optim}.

\subsection{Composed Proposer--Evaluator Architecture}\label{sec:bilevel}

D2C employs a composed proposer-evaluator architecture consisting of a conditional diffusion model $G_\theta$ (the \emph{proposer}) that generates designs from conditioning inputs, and a forward surrogate $S_\phi$ (the \emph{evaluator}) that predicts performance metrics from designs:
\begin{equation}
    \hat{y} = S_\phi(x,\, c_{\mathrm{fix}}), \qquad \hat{y} \in \mathbb{R}^{d_y},
\end{equation}
where $\hat{y}$ are predicted scalar quantities (e.g., energy consumption, yield, carbon emissions).
The surrogate $S_\phi$ is trained independently on the same dataset and remains frozen throughout.

This architecture serves two roles: providing a surrogate consistency signal during training (Section~\ref{sec:training}) and an objective gradient $\nabla_{x_0} \mathcal{L}_{\mathrm{obj}}$ during inference-time optimization (Section~\ref{sec:optim}).
%Because both $G_\theta$ and $S_\phi$ are differentiable, gradients flow end-to-end from the loss through the evaluator and proposer back to the conditioning input.

\subsection{Training the Conditional Diffusion Proposer}\label{sec:training}

The proposer $G_\theta$ is trained offline on a dataset $\mathcal{D} = \{(c_{\mathrm{fix},i},\, x_i,\, y_i)\}_{i=1}^N$, where $y_i \in \mathbb{R}^{d_o}$ are observed performance metrics. The full conditioning is assembled as $c_i = (c_{\mathrm{fix},i},\, y_i)$, so the metrics serve dual roles: as supervised labels for the surrogate $S_\phi$ and as the optimizable conditioning targets $c_{\mathrm{opt}}$ that steer generation.
The training objective combines the standard denoising score-matching loss with auxiliary terms that improve generation fidelity:
\begin{equation}\label{eq:training_loss}
    \mathcal{L}_{\mathrm{train}} = \underbrace{\|\epsilon - \epsilon_\theta(x_t, c, t)\|^2}_{\text{denoising}} + \alpha_r\, \underbrace{\|\hat{x}_0 - x\|^2}_{\text{reconstruction}} + \alpha_a\, \underbrace{\mathcal{L}_{\mathrm{fid}}(\hat{x}_0, x)}_{\text{physics-based fidelity}} + \alpha_s\, \underbrace{\|S_\phi(\hat{x}_0, c_{\mathrm{fix}}) - y\|^2}_{\text{surrogate consistency}},
\end{equation}
where $x_t = \sqrt{\bar\alpha_t}\,x + \sqrt{1-\bar\alpha_t}\,\epsilon$ is the noised sample with $\epsilon \sim \mathcal{N}(0,\mathbf{I})$, and $\hat{x}_0$ is the one-step predicted clean sample:
\begin{equation}
    \hat{x}_0 = \frac{x_t - \sqrt{1 - \bar\alpha_t}\,\epsilon_\theta(x_t, c, t)}{\sqrt{\bar\alpha_t}}.
\end{equation}
The physics-based fidelity term $\mathcal{L}_{\mathrm{fid}}$ encodes domain-specific structural constraints on the generated design, for example, integral conservation laws, peak-amplitude bounds, or spectral regularity.
We express it generically as:
\begin{equation}\label{eq:fidelity}
    \mathcal{L}_{\mathrm{fid}}(\hat{x}_0, x) = \sum_{k} \bigl[g_k(\hat{x}_0) - g_k(x)\bigr]_+,
\end{equation}
where each $g_k$ is a scalar functional encoding a physical constraint and $[\cdot]_+ = \max(\cdot, 0)$ ensures the penalty activates only when the generated design exceeds the reference on that constraint. In the smooth case, the resulting penalty is differentiable almost everywhere; some instantiated constraints additionally use non-smooth operators such as coordinatewise $\max/\min$, which are handled in practice with automatic-differentiation subgradients.
The concrete instantiations of $\mathcal{L}_{\mathrm{fid}}$ and the surrogate consistency term for each application domain are detailed in Appendix~\ref{app:training_losses}.
The surrogate consistency term leverages the composed architecture of Section~\ref{sec:bilevel}: the frozen evaluator $S_\phi$ maps generated designs back to the performance-metric space, encouraging the proposer to produce physically consistent outputs.

To enable classifier-free guidance at inference, the conditioning is randomly replaced with a learned null embedding $\emptyset$ with probability $p_{\mathrm{drop}}$ during training.
Exponential moving averaging (EMA) of model weights with decay $\beta$ further stabilizes generation quality.
Once training converges, all proposer parameters $\theta$ are frozen.

\subsection{Differentiable DDIM Reverse Chain}\label{sec:ddim_chain}

The key observation is that when the DDIM sampler uses zero noise ($\sigma_t = 0$), the entire reverse chain becomes a deterministic, differentiable function of its inputs, including the conditioning $c$.
With the proposer frozen, we establish the differentiable path from $\mathcal{L}$ to $c_{\mathrm{opt}}$ required by Eq.~\eqref{eq:main_opt}.
Each reverse step takes the form:
\begin{equation}\label{eq:ddim_step}
    x_{t-1} = \sqrt{\bar\alpha_{t-1}}\;\hat{x}_0(x_t, c, t) + \sqrt{1 - \bar\alpha_{t-1}}\;\tilde\epsilon_\theta(x_t, c, t),
\end{equation}
where $\hat{x}_0(x_t, c, t) = \bigl(x_t - \sqrt{1 - \bar\alpha_t}\;\tilde\epsilon_\theta\bigr) / \sqrt{\bar\alpha_t}$ is the predicted clean sample, and $\tilde\epsilon_\theta$ is the CFG-guided noise estimate~\citep{ho2022cfg}:
\begin{equation}\label{eq:cfg}
    \tilde\epsilon_\theta(x_t, c, t) = \epsilon_\theta(x_t, \emptyset, t) + w\bigl(\epsilon_\theta(x_t, c, t) - \epsilon_\theta(x_t, \emptyset, t)\bigr),
\end{equation}
with $\emptyset$ the learned null embedding and $w \geq 0$ the guidance scale.
Because every operation is deterministic and differentiable, the full $T$-step composition $G_\theta(c, x_T) = x_0$ is differentiable with respect to~$c$.
The conditioning $c$ is injected at every timestep, providing gradient shortcut connections that mitigate vanishing gradients across deep chains.
The guidance scale $w$ modulates the magnitude of $\partial\tilde\epsilon_\theta / \partial c$, controlling both generation fidelity and optimization stability.

\subsection{Inference-Time Conditioning Optimization}\label{sec:optim}

With $G_\theta$ and $S_\phi$, we optimize $c_{\mathrm{opt}}$ by backpropagating a composite loss through the DDIM chain.
The loss is a weighted combination of $K$ differentiable penalty terms:
\begin{equation}\label{eq:total_loss}
    \mathcal{L} = \sum_{k=1}^{K} \lambda_k\,\mathcal{L}_k(x_0,\, x_{\mathrm{ref}},\, c_{\mathrm{fix}}),
\end{equation}
where each $\mathcal{L}_k$ is a scalar loss computed from the generated design $x_0 = G_\theta(c, x_T)$, the reference design $x_{\mathrm{ref}}$, and/or the fixed conditioning $c_{\mathrm{fix}}$ (e.g., the surrogate evaluation $S_\phi(x_0, c_{\mathrm{fix}})$), and the weights $\boldsymbol\lambda = (\lambda_1, \ldots, \lambda_K)$ control the trade-off among competing objectives.
Crucially, no loss term depends directly on the optimizable component $c_{\mathrm{opt}}$; all losses are functions of the \emph{realized design} $x_0$ only.
The conditioning $c_{\mathrm{opt}}$ influences the loss exclusively through its effect on the generated design via $x_0 = G_\theta((c_{\mathrm{fix}}, c_{\mathrm{opt}}), x_T)$, ensuring that any reported improvement reflects a genuine change in the physical design rather than a shift in target labels.
In the instantiated losses used here, some penalty terms include non-smooth operators such as coordinatewise $\max/\min$, so optimization is implemented through the deterministic reverse chain with practical automatic-differentiation subgradients where needed.

The loss terms are domain-specific but follow a common taxonomy.
A \emph{surrogate objective} term uses the frozen evaluator to maximize a target metric (e.g., $\mathcal{L}_{\mathrm{obj}} = -S_\phi(x_0, c_{\mathrm{fix}})_k$ for target metric $k$); \emph{constraint} terms penalize violations of physical invariants (integral conservation, amplitude bounds, spectral stability); and a \emph{smoothness} term enforces temporal or spatial regularity.
The task-specific loss terms are constructed to be on broadly comparable scales; the concrete loss definitions for each application are given in Appendix~\ref{app:optim_losses}.

At each optimization step, the conditioning is updated via gradient descent:
\begin{equation}\label{eq:cond_update}
    c_{\mathrm{opt}}^{(j+1)} = c_{\mathrm{opt}}^{(j)} - \eta\,\nabla_{c_{\mathrm{opt}}} \mathcal{L}\!\bigl(G_\theta(c^{(j)}, x_T),\, x_{\mathrm{ref}},\, c_{\mathrm{fix}}\bigr),
\end{equation}
where $j = 0, \ldots, J{-}1$ indexes optimization steps, $\eta$ is the learning rate, and $x_T$ is held fixed throughout. In practice, any first-order optimizer (e.g., Adam) can replace vanilla gradient descent; the key requirement is differentiability of the full chain.
Crucially, only $c_{\mathrm{opt}}$ receives gradient updates; $\theta$ and $\phi$ remain frozen.
This mitigates the risk of surrogate exploitation common in unconstrained model-based optimization~\citep{trabucco2021coms}: under suitably tuned guidance, the generative model helps keep designs closer to the empirical data distribution induced by the learned prior while the surrogate provides only a directional gradient signal.

\subsection{Stochastic Coverage and Candidate Selection}\label{sec:coverage}

Since the mapping $G_\theta(c, x_T)$ depends on the latent noise $x_T$, a single noise draw may not expose the best mode of the conditional distribution.
To ensure broad coverage without sacrificing differentiability, we adopt a two-level diversification strategy.

\paragraph{Noise trials.}
For each input, we draw $N_z$ independent noise vectors $\{x_T^{(n)}\}_{n=1}^{N_z}$, each held fixed during optimization.
Each trial explores a different region of the latent space, producing structurally distinct candidates.

\paragraph{Weight diversity.}
Within each noise trial, we instantiate $N_\lambda$ loss-weight vectors $\boldsymbol\lambda^{(m)} \sim \mathrm{Dir}(\boldsymbol\alpha)$ drawn from a symmetric Dirichlet distribution~\citep{frigyik2010dirichlet} with $\alpha_k = 1$ for all $k$ (i.e., uniform over the simplex; see Appendix~\ref{app:dirichlet} for details).
The Dirichlet distribution ensures each weight vector sums to one and covers the full trade-off simplex uniformly, so no single objective dominates across all trajectories.
This yields $B = N_z \times N_\lambda$ parallel trajectories per sample, each emphasizing a different trade-off among competing objectives.

\paragraph{Candidate selection.} \label{candidate_validation}
After $J$ optimization steps, each trajectory produces a candidate $x_0^{(b)}$.
Each candidate is then validated against domain-specific hard physical constraints $\mathcal{C}(x_0) \leq \mathbf{0}$ (e.g., temperature limits, integral tolerances; see Appendix~\ref{sec:domain_specific_filtering} for details).
Among candidates that satisfy the hard constraints, the one with the best primary objective is selected; if no candidate passes all constraints, the baseline design $x_{\mathrm{ref}}$ is retained (yielding $0\%$ gain).

\begin{algorithm}[hptb]
\caption{D2C: Differentiable Diffusion Conditioning}
\label{alg:d2c}
\begin{algorithmic}[1]
\REQUIRE Dataset $\mathcal{D} = \{(c_{\mathrm{fix},i},\, x_i,\, y_i)\}$ with $c_i = (c_{\mathrm{fix},i},\, y_i)$; test sample $(c,\, x_{\mathrm{ref}})$
\REQUIRE Noise trials $N_z$, weight samples $N_\lambda$, optimization steps $J$, guidance scale $w$
%\STATE \textbf{Stage 1: Offline Training} (performed once)
\STATE Train surrogate $S_\phi$ on $\{(x_i,\, c_{\mathrm{fix},i}) \to y_i\}$; freeze $\phi$
\STATE Train $\epsilon_\theta$ on $\mathcal{D}$ using $\mathcal{L}_{\mathrm{train}}$ (Eq.~\ref{eq:training_loss}) with CFG dropout $p_{\mathrm{drop}}$ and frozen $S_\phi$; freeze $\theta$
\STATE
%\STATE \textbf{Stage 2: Conditioning Optimization} (per test sample)
\STATE Decompose $c = (c_{\mathrm{fix}},\, c_{\mathrm{opt}}^{(0)})$
\STATE $\mathcal{R} \leftarrow \emptyset$ \COMMENT{candidate pool}
\FOR{$n = 1, \ldots, N_z$}
    \STATE Sample $x_T^{(n)} \sim \mathcal{N}(0, \mathbf{I})$
    \FOR{$m = 1, \ldots, N_\lambda$}
        \STATE Sample $\boldsymbol\lambda^{(m)} \sim \mathrm{Dir}(\boldsymbol\alpha)$
        \STATE Initialize $c_{\mathrm{opt}} \leftarrow c_{\mathrm{opt}}^{(0)}$
        \FOR{$j = 1, \ldots, J$}
            \STATE $x_0 \leftarrow G_\theta\bigl((c_{\mathrm{fix}},\, c_{\mathrm{opt}}),\, x_T^{(n)};\, w\bigr)$ \COMMENT{Eqs.~\ref{eq:ddim_step}--\ref{eq:cfg}}
            \STATE $\mathcal{L} \leftarrow \sum_{k} \lambda_k^{(m)}\,\mathcal{L}_k(x_0,\, x_{\mathrm{ref}},\, c_{\mathrm{fix}})$ \COMMENT{Eq.~\ref{eq:total_loss}}
            \STATE $c_{\mathrm{opt}} \leftarrow c_{\mathrm{opt}} - \eta\,\nabla_{c_{\mathrm{opt}}} \mathcal{L}$ \COMMENT{backprop through $G_\theta$, $S_\phi$}
        \ENDFOR
        \STATE $\mathcal{R} \leftarrow \mathcal{R} \cup \bigl\{G_\theta\bigl((c_{\mathrm{fix}},\, c_{\mathrm{opt}}),\, x_T^{(n)};\, w\bigr)\bigr\}$
    \ENDFOR
\ENDFOR
\STATE
%\STATE \textbf{Stage 3: Validity-First Selection}
\STATE $\mathcal{V} \leftarrow \{x \in \mathcal{R} \mid \mathcal{C}(x) \leq \mathbf{0}\}$ \COMMENT{filter by hard constraints}
\IF{$\mathcal{V} \neq \emptyset$}
    \STATE \textbf{return} $\arg\min_{x \in \mathcal{V}} \mathcal{L}_{\mathrm{obj}}(x)$ \COMMENT{surrogate objective}
\ELSE
    \STATE \textbf{return} $x_{\mathrm{ref}}$ \COMMENT{no valid improving candidate; retain baseline (0\% gain)}
\ENDIF
\end{algorithmic}
\end{algorithm}

%Algorithm~\ref{alg:d2c} summarizes D2C. Stage~1 trains the evaluator and proposer offline, Stage~2 performs batched gradient-based conditioning optimization over $B = N_z \times N_\lambda$ trajectories per sample, and Stage~3 selects the best feasible candidate.

% ─────────────────────────────────────────────────────────────
\section{Experiments}\label{sec:experiments}
% ─────────────────────────────────────────────────────────────

We evaluate D2C on two sets of tasks: (i)~two real-world physical design problems where constraint satisfaction is safety-critical, and (ii)~the standard Design-Bench~\citep{trabucco2022designbench} benchmark suite for offline model-based optimization. Although D2C is motivated by constrained physical domains, we also evaluate it on Design-Bench to test whether the same mechanism remains strong on standard offline MBO benchmarks beyond the application domains that motivate the method.
Additional implementation, hyperparameter, and evaluation details are provided in Appendices~\ref{app:exp_details} and~\ref{app:designbench_data}.

\subsection{Real-World Physical Design Tasks}\label{sec:real_tasks}

\paragraph{Inertial Confinement Fusion (ICF) laser pulse optimization.}
In ICF, a precisely shaped laser pulse drives the implosion of a deuterium-tritium fuel capsule to achieve thermonuclear ignition. The design variable $x \in \mathbb{R}^{256}$ is a temporal laser power profile discretized into 256 time steps, and the conditioning signal $c$ encodes target specifications ($c_{\mathrm{fix}}$, 7 dimensions) and 20 performance metrics ($c_{\mathrm{opt}}$) derived from LILAC, the radiation-hydrodynamics simulation code used for inertial-confinement-fusion design studies. At inference, D2C optimizes the 20 performance targets to steer the proposer toward higher-yield pulses. The primary objective is to maximize neutron yield, subject to constraints on pulse energy conservation (the integral of the optimized pulse must match the reference), temporal smoothness (to ensure manufacturability), and spectral fidelity (the frequency content must remain physically plausible). Designs that violate these constraints are operationally unsafe and cannot be fielded on a real facility; the concrete validation criteria are detailed in Appendix~\ref{sec:domain_specific_filtering}.

\paragraph{Data center workload scheduling.}
For this task we follow the SustainDC \citep{sustaindc2024} framework where the goal is to reduce the carbon footprint of large-scale data centers, which requires scheduling computational workloads to exploit temporal variations in grid carbon intensity and cooling efficiency. Here, $x \in \mathbb{R}^{144}$ is a 24-hour workload power profile at 10-minute resolution, and $c$ encodes environmental state variables ($c_{\mathrm{fix}}$, 73 dimensions: ambient temperatures, cooling set-points, IT demand) and energy metrics ($c_{\mathrm{opt}}$, 10 dimensions: total power, carbon intensity, etc.). At inference, D2C optimizes the 10 energy metric targets to minimize emissions. The objective is to minimize operational carbon emissions while satisfying a hard workload conservation constraint: the total compute delivered must equal the demand (the integral of the optimized profile must match the reference within a tight tolerance). Additional constraints include inlet temperature safety limits. Final designs are validated against the full SustainDC thermal model (Appendix~\ref{sec:domain_specific_filtering}).

\subsection{Design-Bench Benchmarks}\label{sec:designbench}

To position D2C among established MBO methods, we evaluate on four tasks from Design-Bench~\citep{trabucco2022designbench}:  \textbf{Ant Morphology} (continuous, 60-d), \textbf{D'Kitty Morphology} (continuous, 56-d), \textbf{TF Bind 8} (discrete, 8-d), and \textbf{TF Bind 10} (discrete, 10-d). These tasks span transcription factor binding (DNA sequence design) and robot morphology optimization, providing a diverse testbed for offline optimization from fixed datasets. The goal in each task is to maximize a black-box oracle score using only the offline dataset.
\subsection{Baselines}\label{sec:baselines}

We compare against 12 baselines, 11 of which are commonly used in offline MBO:
Gradient Ascent, COMs~\citep{trabucco2021coms}, BDI~\citep{chen2022bdi}, CbAS~\citep{brookes2019cbas}, MINs~\citep{kumar2020mins}, REINFORCE~\citep{williams1992reinforce}, DDOM~\citep{krishnamoorthy2023ddom}, GTG~\citep{yun2024gtg}, PGS~\citep{chemingui2024pgs}, GP-qEI~\citep{snoek2012practical}, CMA-ES~\citep{hansen2016cmaes}, and Brute Force, a diffusion-based approach that replaces gradient-based conditioning optimization with a forward-only sweep over $c_{\mathrm{opt}}$ (Appendix~\ref{app:exp_details}).
For the real-world tasks, all surrogate-based baselines share the same frozen $S_\phi$.

\subsection{Main Results}\label{sec:results}

Table~\ref{tab:real_results} reports results on the real-world tasks.
All results report mean $\pm$ standard deviation over 5 independent runs with different random seeds; the same seeds and test samples are used across all methods.
Following the standard MBO candidate-generation protocol~\citep{trabucco2022designbench, krishnamoorthy2023ddom, yun2024gtg}, all baselines generate 128 candidate designs per test sample; D2C generates 100 candidates per sample. For the real-world tasks, these 100 candidates arise from $N_z{=}20$ noise trials $\times$ $N_\lambda{=}5$ weight samples; for Design-Bench, we use 100 latent/noise trials ($N_\lambda{=}1$ since single objective) with the Design-Bench-specific setup described in Appendix~\ref{app:designbench_data}. For the real-world tasks we report gains in terms of \% neutron yield improvement for ICF and \% carbon-footprint reduction for the data center task. For ICF, final performance is evaluated by a separate LILAC-validated evaluation oracle (Appendix~\ref{app:icf_oracle}); for DC, generated workloads are evaluated directly on the SustainDC physics-based simulator. For Design-Bench, we report 100th-percentile normalized score of the best design found as is standard~\citep{trabucco2022designbench, krishnamoorthy2023ddom, yun2024gtg}.
ICF Standard evaluates optimization over 2{,}000 randomly sampled pulses from the test set (the same 2{,}000 for all methods); ICF Top selects the 1{,}000 highest-yield designs from the test set and optimizes only those, stress-testing each method on the hardest cases where the baseline design is already among the best.
For all tasks, only candidates that pass the domain-specific hard constraints (Section~\ref{candidate_validation}) are counted as improvements; samples where no valid improving candidate is found retain the baseline design and contribute $0\%$ gain.

\begin{table}[t]
\caption{Real-world physical design tasks. Values are mean percentage improvement (\%) over the baseline design $\pm$ std over 5 runs. \textbf{Bold}: best.}
\label{tab:real_results}
\centering
\small
\setlength{\tabcolsep}{4pt}
\begin{tabular}{lccc}
\toprule
Method & ICF Standard & ICF Top & DC \\
\midrule
Gradient Ascent  & $2.31\% \pm 0.31$  & $1.46\% \pm 0.29$  & $0.14\% \pm 0.04$ \\
COMs             & $3.14\% \pm 0.41$  & $0.43\% \pm 0.11$  & $1.33\% \pm 0.24$ \\
BDI              & $2.42\% \pm 0.37$  & $1.33\% \pm 0.21$  & $0.37\% \pm 0.08$ \\
CbAS             & $4.49\% \pm 0.63$  & $1.86\% \pm 0.29$  & $1.85\% \pm 0.31$ \\
MINs             & $3.71\% \pm 0.67$  & $1.51\% \pm 0.43$  & $1.21\% \pm 0.23$ \\
REINFORCE        & $3.50\% \pm 0.56$  & $0.32\% \pm 0.08$  & $1.26\% \pm 0.25$ \\
DDOM             & $4.55\% \pm 0.68$ & $2.24\% \pm 0.25$  & $4.67\% \pm 0.59$ \\
GTG              & $5.43\% \pm 0.74$  & $2.65\% \pm 0.38$  & $5.50\% \pm 0.68$ \\
PGS              & $2.47\% \pm 0.42$ & $1.73\% \pm 0.27$  & $0.42\% \pm 0.10$ \\
GP-qEI           & $3.58\% \pm 0.53$  & $1.03\% \pm 0.18$  & $5.93\% \pm 0.82$ \\
CMA-ES           & $4.13\% \pm 0.79$  & $1.24\% \pm 0.27$  & $3.98\% \pm 0.63$ \\
Brute Force      & $4.66\% \pm 0.83$ & $3.93\% \pm 0.52$  & $1.14\% \pm 0.22$ \\
\midrule
D2C (Ours)      & $\mathbf{17.98\% \pm 1.54}$ & $\mathbf{10.06\% \pm 0.88}$ & $\mathbf{8.98\% \pm 0.75}$ \\
\bottomrule
\end{tabular}
\end{table}

\paragraph{Real-world tasks.}
D2C achieves the highest improvement on all three real-world settings (Table~\ref{tab:real_results}).
On ICF Standard, D2C improves yield by $17.98\%$, outperforming all baselines by a large margin.
On ICF Top, where the baseline designs are already high-yield, D2C achieves $10.06\%$, more than two and a half times the best baseline, Brute Force ($3.93\%$).
%Beyond this large-scale proxy evaluation, a separate small-scale LILAC validation on 20 expert-designed pulses showed an average yield gain of $9.8\%$ over the original expert baselines (Appendix~\ref{app:icf_oracle}).
This suggests that differentiable guidance through the DDIM chain enables more precise conditioning than fixed-target generation, especially in the high-performance regime.
On data center scheduling (DC), D2C reduces carbon emissions by $8.98\%$, surpassing the best baseline GP-qEI ($5.93\%$) and the best generative baseline GTG ($5.50\%$).

\begin{table}[t]
\caption{Design-Bench tasks. 100th-percentile normalized score $\pm$ std over 5 runs. The row \emph{$\mathcal{D}$ (best)} reports the best design present in the offline dataset for each task. Scores above \emph{$\mathcal{D}$ (best)} therefore indicate improvement beyond the best offline example under the evaluation protocol used for that column. Avg.\ Rank is the mean ordinal rank across all four tasks ($\downarrow$ lower is better). \textbf{Bold}: best per column.}
\label{tab:designbench_results}
\centering
\small
\setlength{\tabcolsep}{4pt}
\begin{tabular}{lccccc}
\toprule
Method & TF Bind 8 & TF Bind 10 & D'Kitty & Ant & Avg.\ Rank \\
\midrule
$\mathcal{D}$ (best) & $0.439$ & $0.467$ & $0.884$ & $0.565$ & -- \\
Gradient Ascent & $0.854 \pm 0.029$ & $0.575 \pm 0.039$ & $0.937 \pm 0.014$ & $0.749 \pm 0.034$ & $10.38$ \\
COMs            & $0.949 \pm 0.018$ & $0.598 \pm 0.034$ & $0.935 \pm 0.013$ & $0.863 \pm 0.021$ & $6.75$ \\
BDI             & $0.969 \pm 0.012$ & $0.609 \pm 0.026$ & $0.931 \pm 0.014$ & $0.717 \pm 0.032$ & $7.00$ \\
CbAS            & $0.918 \pm 0.027$ & $0.607 \pm 0.032$ & $0.908 \pm 0.017$ & $0.672 \pm 0.035$ & $11.25$ \\
MINs            & $0.923 \pm 0.026$ & $0.625 \pm 0.033$ & $0.917 \pm 0.017$ & $0.806 \pm 0.030$ & $8.50$ \\
REINFORCE       & $0.949 \pm 0.022$ & $0.594 \pm 0.035$ & $0.941 \pm 0.013$ & $0.852 \pm 0.028$ & $7.00$ \\
DDOM            & $0.964 \pm 0.015$ & $0.682 \pm 0.025$ & $0.915 \pm 0.016$ & $\mathbf{0.925 \pm 0.019}$ & $4.38$ \\
GTG             & $0.949 \pm 0.020$ & $0.635 \pm 0.030$ & $\mathbf{0.976 \pm 0.009}$ & $0.905 \pm 0.022$ & $3.50$ \\
PGS             & $0.949 \pm 0.019$ & $0.575 \pm 0.037$ & $0.915 \pm 0.016$ & $0.853 \pm 0.027$ & $9.00$ \\
GP-qEI          & $0.924 \pm 0.024$ & $0.583 \pm 0.036$ & $0.948 \pm 0.012$ & $0.801 \pm 0.029$ & $8.50$ \\
CMA-ES          & $0.949 \pm 0.019$ & $0.638 \pm 0.019$ & $0.929 \pm 0.015$ & $0.843 \pm 0.023$ & $6.50$ \\
Brute Force     & $0.968 \pm 0.014$ & $0.611 \pm 0.028$ & $0.888 \pm 0.019$ & $0.874 \pm 0.026$ & $6.50$ \\
\midrule
D2C (Ours)     & $\mathbf{0.973 \pm 0.011}$ & $\mathbf{0.684 \pm 0.023}$ & $0.974 \pm 0.009$ & $0.876 \pm 0.020$ & $\mathbf{1.75}$ \\
\bottomrule
\end{tabular}
\end{table}

\paragraph{Design-Bench tasks.}
D2C achieves the best average rank across the four Design-Bench tasks, placing first on TF Bind 8 and TF Bind 10, second on D'Kitty, and third on Ant. We therefore position D2C as robust across standard MBO benchmarks, with the strongest overall performance by average rank.

Additional ablations on guidance scale (including its selection without oracle access), optimization steps, learning rate, DDIM steps, optimizable conditioning dimensionality, conditioning versus latent-noise optimization, constraint-satisfying improvement rate, candidate budget, and computational cost are reported in Section~\ref{sec:ablations}. In particular, with the proposer, surrogate, and evaluation protocol held fixed, optimizing the conditioning outperforms optimizing the latent noise on both real-world tasks (Section~\ref{sec:ablation_cond_vs_latent}).

% ─────────────────────────────────────────────────────────────
\section{Related Work}
% ─────────────────────────────────────────────────────────────

Offline model-based optimization (MBO) seeks high-scoring designs using only a fixed dataset, with no online evaluations~\citep{trabucco2022designbench}. Existing approaches include forward methods that optimize a learned surrogate in design space~\citep{trabucco2021coms, fu2021nemo, yu2021roma, yuan2023ict, chen2023parallel}, inverse methods that map target scores to designs with increasingly expressive generators~\citep{brookes2019cbas, fannjiang2020autofocused, kumar2020mins, kim2023bootstrapped, krishnamoorthy2023ddom}, and trajectory-based methods that learn to improve from synthetic optimization paths~\citep{mashkaria2023bonet, chemingui2024pgs, yun2024gtg}. Forward methods remain limited by surrogate extrapolation, while inverse and trajectory-based methods typically keep the conditioning target fixed at inference time rather than refining it against downstream objectives.

Recent diffusion-based optimization methods attempt to integrate optimization feedback into generation, including guided warm-starts with Langevin corrections~\citep{kong2024diffopt}, uncertainty-aware conditioning selection~\citep{wu2024diffbbo}, editing-based optimization against pseudo-targets~\citep{yuan2024demo}, and end-to-end gradient propagation through diffusion samplers in image-generation settings~\citep{wallace2023doodl, pan2024adjointdpm}. However, these methods either operate outside offline MBO, rely on post-hoc corrections or heuristic target selection, or do not study physics-informed constrained design with a frozen proposer-evaluator architecture. Unlike these methods, D2C targets offline MBO and treats the deterministic reverse diffusion chain as a differentiable map in the conditioning signal itself, optimizing that conditioning directly against surrogate and physics-based penalty losses, with final validity enforced by domain-specific hard constraints.

\section{Conclusion and Limitations}
% ─────────────────────────────────────────────────────────────

We introduced Differentiable Diffusion Conditioning (D2C), a framework that turns a pretrained conditional diffusion model into an end-to-end gradient-based optimizer for physical design. By establishing a practical backpropagation path from composite physics-informed objectives, through a frozen surrogate evaluator, back through the deterministic DDIM reverse chain to the conditioning input, D2C enables gradient-based optimization without retraining or fine-tuning the generative model. The composed proposer-evaluator architecture of D2C helps keep generated designs closer to the empirical data distribution induced by the learned prior under appropriately tuned guidance while the surrogate provides directional gradient signals, helping reduce the out-of-distribution exploitation that plagues direct surrogate optimization. Across two real-world scientific tasks and four Design-Bench tasks, D2C delivers the strongest results on the real-world tasks and the best average rank across the benchmark suite.

\subsection{Limitations}

\begin{itemize}[leftmargin=12pt]
    \item Computational cost. D2C requires backpropagation through a $T$-step DDIM chain at each optimization iteration, making it roughly $8$--$17\times$ slower than single-pass methods (Table~\ref{tab:comp_cost}). While our ablations show that $T$ and $J$ can be reduced with minimal quality loss, the method remains unsuitable for latency-critical applications.
    \item Surrogate dependence. The quality of the gradient signal is bounded by the accuracy of the frozen surrogate $S_\phi$. In regimes where the surrogate is poorly calibrated, the optimization may converge to high-scoring surrogate designs that perform poorly on the true objective. Future work could incorporate uncertainty quantification into the surrogate to mitigate this.
    \item Validity-first selection. D2C combines differentiable penalties with a final validity-first selection step. As a result, method quality should be interpreted jointly through both achieved gain and the rate at which valid improving solutions are found.
    \item Domain-specific tuning. While D2C is broadly applicable, the choice of loss terms, constraint definitions, and optimization hyperparameters requires domain expertise and may involve trial-and-error tuning for new applications. Appendix~\ref{app:practical_guidance} provides extended practical guidance.
\end{itemize}

\section*{Acknowledgments} This work was supported by the U.S. Department of Energy, Office of Science, under Award No. DE-SC0024408 and DE-SC0024381.

\bibliographystyle{plainnat}
\bibliography{ref}

%%%%%%%%%%%%%%%%%%%%%%%%%%%%%%%%%%%%%%%%%%%%%%%%%%%%%%%%%%%%

\newpage
\appendix

\section{Experimental Details}\label{app:exp_details}

\subsection{Computational Resources}

All experiments were conducted on a node equipped with four NVIDIA H100 GPUs. Training the diffusion proposer and the surrogate evaluator together on a single GPU takes approximately 20 hours per training run. For the optimization time on this same hardware please refer to Subsection~\ref{sec:ablation_cost}.

\subsection{Diffusion Proposer Architecture}

The proposer $G_\theta$ is an MLP-based epsilon-prediction network.
Both ICF and DC tasks use the same architecture design; only the input/output dimensionalities and hidden width differ.
The architecture consists of:
\begin{itemize}[leftmargin=2em,itemsep=1pt]
    \item A \textbf{data projection} layer: $\texttt{Linear}(d_{\mathrm{out}}, H) \to \texttt{LayerNorm} \to \texttt{SiLU}$, where $d_{\mathrm{out}}$ is the design dimensionality and $H$ is the hidden dimension.
    \item A \textbf{time embedding} MLP: sinusoidal positional encoding of the diffusion timestep, followed by three linear layers with SiLU activations, projecting to dimension $H$.
    \item A \textbf{conditioning MLP}: three linear layers with SiLU activations mapping the full conditioning vector $c \in \mathbb{R}^{d_c}$ to dimension $H$.
    \item $L$ \textbf{residual blocks}, each containing LayerNorm $\to$ Linear$(H, 4H)$ $\to$ SiLU $\to$ Dropout $\to$ Linear$(4H, H)$ $\to$ Dropout, with a learnable residual scale parameter initialized to $0.1$.
    \item Per-block \textbf{FiLM modulation}: at each residual block, the time embedding produces scale and shift vectors via dedicated linear projections, applied as $h \leftarrow \gamma_t \odot h + \beta_t$. The conditioning embedding is added to the hidden state before the residual stack.
    \item A \textbf{null conditioning embedding} (learnable, $\mathbb{R}^{1 \times H}$, initialized to $\mathcal{N}(0, 0.01)$) for classifier-free guidance: during training, the conditioning embedding is replaced by the null embedding with probability $p_{\mathrm{drop}}$.
    \item An \textbf{output projection}: LayerNorm $\to$ Linear$(H, d_{\mathrm{out}})$, with weights and biases initialized to zero for stable early training.
\end{itemize}

\subsection{Noise Schedule}

Both tasks use a cosine noise schedule~\citep{nichol2021improved} with $T_{\mathrm{train}} = 100$ diffusion steps, $s = 0.008$ offset, and betas clipped to $[10^{-4},\, 0.999]$:
\begin{equation}
    \bar\alpha_t = \frac{f(t)}{f(0)}, \quad f(t) = \cos\!\left(\frac{t/T + s}{1 + s} \cdot \frac{\pi}{2}\right)^2.
\end{equation}
At inference, we use DDIM with $\sigma_t = 0$ (fully deterministic) and $T = 25$ inference steps (selected via ablation, Section~\ref{sec:ablation_inference}).

\subsection{Training Hyperparameters}

The proposer is trained with AdamW~\citep{loshchilov2019decoupled} using a cosine-annealed learning rate with linear warmup.
EMA (Exponential Moving Average) of model weights is maintained throughout training with a warmup period of 2{,}000 steps.
Table~\ref{tab:training_hparams} summarizes the training hyperparameters for both domains.

\begin{table}[ht]
\caption{Training hyperparameters for the diffusion proposer.}
\label{tab:training_hparams}
\centering
\small
\begin{tabular}{lcc}
\toprule
Hyperparameter & ICF & DC \\
\midrule
Hidden dimension $H$ & 1024 & 1024 \\
Residual blocks $L$ & 6 & 6 \\
Dropout & 0.1 & 0.1 \\
CFG dropout $p_{\mathrm{drop}}$ & 0.1 & 0.1 \\
EMA decay $\beta$ & 0.9999 & 0.9999 \\
Learning rate & $5 \times 10^{-5}$ & $5 \times 10^{-5}$ \\
Batch size & 1024 & 1024 \\
Training epochs & 400 & 5{,}000 \\
Diffusion steps $T_{\mathrm{train}}$ & 100 & 100 \\
Noise schedule & Cosine & Cosine \\
Reconstruction weight $\alpha_r$ & 0.5 & 0.5 \\
Fidelity weight $\alpha_a$ & 0.001 & 0.01 \\
Smoothness weight $\alpha_{\mathrm{sm}}$ & 0.0001 & 0.0001 \\
Surrogate weight $\alpha_s$ & 1.0 & 1.0 \\
Conditioning dim $d_c$ & 27 & 83 \\
Design dim $d_{\mathrm{out}}$ & 256 & 144 \\
\bottomrule
\end{tabular}
\end{table}

\subsection{Inference-Time Optimization Hyperparameters}

At inference, we freeze the trained proposer and evaluator and optimize the conditioning separately for each test sample.
Table~\ref{tab:inference_hparams} reports the optimization hyperparameters used for the main results (selected via ablation studies in Section~\ref{sec:ablations}).

Hyperparameters were selected via grid search on the \emph{validation} set over manually specified value ranges.
During inference-time optimization, only the optimizable coordinates of the conditioning vector are updated; the fixed coordinates remain unchanged.
After each Adam step, the updated conditioning coordinates are clipped back to the normalized $[0,1]$ range before the next DDIM rollout, so the frozen proposer is not queried with out-of-range values on those dimensions.

\begin{table}[ht]
\caption{Inference-time optimization hyperparameters.}
\label{tab:inference_hparams}
\centering
\small
\begin{tabular}{lcc}
\toprule
Hyperparameter & ICF & DC \\
\midrule
Guidance scale $w$ & 2.0 & 2.0 \\
DDIM inference steps $T$ & 25 & 25 \\
Optimization steps $J$ & 100 & 40 \\
Learning rate $\eta$ & $1 \times 10^{-3}$ & $2 \times 10^{-2}$ \\
Noise trials $N_z$ & 20 & 20 \\
Weight samples $N_\lambda$ per trial & 5 & 5 \\
Total candidates $B$ per sample & 100 & 100 \\
Optimizable conditioning dims $d_o$ & 20 (LILAC variables) & 10 (energy metrics) \\
Optimizer & Adam & Adam \\
Barrier coefficient $\mu$ (DC workload loss) & --- & 100 \\
\midrule
\multicolumn{3}{l}{\textit{Validity thresholds}} \\
Minimum gain $\delta_{\mathrm{gain}}$ & $0.1\%$ & $0.1\%$ \\
Spectral tolerance $\tau_{\mathrm{fft}}$ & 0.2 & --- \\
Correlation threshold $\tau_{\mathrm{corr}}$ & 0.8 & --- \\
Workload deviation tolerance & --- & $1\%$ \\
\bottomrule
\end{tabular}
\end{table}

\subsection{Baselines Implementations and Hyperparameters}

For the \textbf{Design-Bench} tasks, baseline implementations and hyperparameters are sourced from the official \texttt{design-baselines} repository\footnote{\url{https://github.com/brandontrabucco/design-baselines}}, the GTG authors' implementation\footnote{\url{https://github.com/dbsxodud-11/GTG}}, and the DDOM authors' implementation\footnote{\url{https://github.com/siddarthk97/ddom}}.
All baselines use the hyperparameters from their respective repositories. 

For the \textbf{real-world tasks}, we implement all baselines that require a surrogate model using the same frozen $S_\phi$ to ensure a fair comparison. For approaches that require training a neural network or diffusion model for test-time optimization, we use the same hidden dimension as our proposer when applicable.
For gradient-based baselines, we perform grid search over learning rates $\{5 \times 10^{-4}, 1 \times 10^{-3}, 5 \times 10^{-3}, 1 \times 10^{-2}, 5 \times 10^{-2}\}$ and select the best-performing value for each method and task. For the evolutionary baseline CMA-ES, we use the \texttt{cma} Python package with default hyperparameters except for population size. For GP-qEI, we use a Gaussian Process surrogate with a Mat\'ern kernel ($\nu=2.5$). %and select candidates via analytical expected improvement over random perturbations.

For the diffusion-based \textbf{Brute Force} baseline on the real-world tasks, we do not perform a single one-shot DDIM sample. Instead, we reuse the same frozen proposer as D2C but replace gradient-based conditioning updates with an explicit sweep over hand-chosen conditioning directions. On ICF, the baseline linearly increases the yield-related conditioning coordinate and runs one full DDIM rollout at each sweep point, retaining the best candidate that passes the same gain, area, FFT, and correlation checks used for D2C. On DC, it linearly decreases selected energy-metric conditioning coordinates and runs repeated DDIM rollouts for multiple noise seeds at each sweep point, followed by the same workload-conservation and inlet-temperature checks used for D2C. Thus, Brute Force should be interpreted as a forward-only conditioning sweep with many repeated DDIM generations, not as a single batched generation pass.

For the surrogate-optimized real-world baselines, we do not optimize the primary objective alone. To reduce the chance that differences in constraint-satisfying improvement rate are an artifact of objective-only optimization, Gradient Ascent, REINFORCE, and CMA-ES are augmented with domain-aligned penalties or projections that mirror the main constraint families used by D2C, although these baselines operate directly in design space or in low-dimensional latent parameterizations rather than through the conditioning optimization of Eq.~\ref{eq:total_loss}. On ICF, Gradient Ascent maximizes surrogate-predicted yield while adding similarity and first/second-order smoothness regularization, REINFORCE uses reward terms for yield together with area and spectral penalties, and CMA-ES searches in a PCA-reduced pulse space with an objective that combines surrogate-predicted yield with similarity, area, and squared first-difference penalties, together with a hard projection that prevents pulse area from exceeding the reference; all three are then validated using the same gain, area, FFT, and correlation thresholds as D2C. On DC, Gradient Ascent minimizes surrogate-predicted carbon emissions with area-conservation and similarity penalties, REINFORCE uses carbon-reduction reward together with an area-conservation penalty, and CMA-ES searches in a PCA-reduced profile space with carbon-reduction objective plus area-conservation, similarity, and squared first-difference penalties; all three are then filtered by the same workload-conservation and inlet-temperature constraints used for D2C. Thus, the lower constraint-satisfying improvement rates of these baselines in Table~\ref{tab:ablation_validity} should not be interpreted as evidence that they ignore constraint-aware optimization entirely.

For all real-world baselines, hyperparameters were selected on the validation split only, and the test split was used exclusively for final reporting.

\subsection{Surrogate Evaluator Architecture}

The frozen evaluator $S_\phi$ is a feedforward residual network with LayerNorm and dropout, trained independently via supervised regression.
For ICF, the surrogate has 10 hidden layers of width 1024 with residual connections every two layers, predicting 20 LILAC performance variables from a 263-dimensional input (256-d pulse + 7 target parameters).
For DC, the surrogate has 6 hidden layers of width 1024 with residual connections every two layers, predicting 10 energy metrics from a 217-dimensional input (144-d workload profile + 73 environmental conditions).
Both surrogates use ReLU activations, LayerNorm, 10\% dropout during training, and a sigmoid output layer mapping to $[0, 1]$ (targets are MinMax-scaled).

Table~\ref{tab:surrogate_hparams} summarizes the training hyperparameters for the surrogate evaluator.

\begin{table}[ht]
\caption{Training hyperparameters for the surrogate evaluator $S_\phi$.}
\label{tab:surrogate_hparams}
\centering
\small
\begin{tabular}{lcc}
\toprule
Hyperparameter & ICF & DC \\
\midrule
Hidden dimension & 1024 & 1024 \\
Hidden layers & 10 & 6 \\
Residual connections & Every 2 layers & Every 2 layers \\
Activation & ReLU & ReLU \\
Dropout (training) & 0.1 & 0.1 \\
Output activation & Sigmoid & Sigmoid \\
Optimizer & AdamW & AdamW \\
Learning rate & $1 \times 10^{-3}$ & $1 \times 10^{-3}$ \\
Weight decay & $1 \times 10^{-5}$ & $1 \times 10^{-5}$ \\
LR schedule & Cosine warm restarts & Cosine warm restarts \\
Batch size & 1024 & 1024 \\
Early stopping patience & 5 (checked every 100 epochs) & 10 (checked every 100 epochs) \\
Gradient clipping & 1.0 & 1.0 \\
Input dimension & 263 (256 + 7) & 217 (144 + 73) \\
Output dimension & 20 & 10 \\
\bottomrule
\end{tabular}
\end{table}

\section{Data and Evaluation}

\subsection{ICF}
The training dataset comprises ${\sim}2{,}600{,}000$ radiation-hydrodynamics simulations from the LILAC simulation database.
Each sample consists of a 256-dimensional laser pulse profile and 27 conditioning dimensions (7 target parameters + 20 LILAC performance variables).
The data are split into 80\% training, 10\% validation, and 10\% test (${\sim}260{,}000$ samples each). The validation set was used for hyperparameter tuning; the test set is used exclusively for final evaluation. We randomly sample 2{,}000 pulses from the test set and use the same 2{,}000 for all methods.
ICF Top selects the 1{,}000 highest-yield samples from the test set.
The surrogate $S_\phi$ achieves $<1\%$ WMAPE on all 20 predicted LILAC variables.

\subsubsection{ICF Data Generation}

The ICF dataset is generated from radiation-hydrodynamics simulations (LILAC code), where each simulation is specified by a target design and a laser pulse shape. LILAC is a prominent one-dimensional Lagrangian hydrodynamics ICF code~\citep{Hansen_2018, LLEReview85, PhysRevA.36.3926, PhysRevLett, Gopalaswamy2019-tc, Gopalaswamy2024-yn}. 
The laser pulse shape is prescribed via an internal library that parameterizes it into descriptive features such as inter-pulse time delays between Gaussian-shaped components, maximum power, and total energy.
The full input specification thus lives in a 27-dimensional parameter space consisting of target specifications and a parameterized laser-pulse description.

Although this 27-D space permits uniform sampling subject to physical constraints (e.g., maximum laser energy, allowable target thicknesses), such sampling does not reflect how experiments or simulations are actually designed.
In practice, high-performing implosions occupy only a small region of this high-dimensional space, guided by physics intent rather than being distributed uniformly across the full admissible hypervolume.
The region of desirable inputs therefore lies on a lower-dimensional manifold embedded within the 27-D space, a manifold that is not explicitly known or parameterized but emerges from complex physics couplings between target geometry and laser-pulse features.

To sample this manifold more effectively, we supplement random draws with perturbations of known high-performing implosion designs.
Perturbing their laser-pulse features and target parameters produces a family of nearby points that remain close to the underlying physics manifold while still providing variation in implosion performance.
This strategy captures the empirical observation that, for a fixed target specification, laser-pulse shapes close to physics-informed designs typically outperform large random perturbations.
The resulting dataset therefore reflects a distribution with multiple high-performance clusters concentrated on an unknown manifold within the 27-D input space, rather than a uniform distribution over the full hypervolume.

\noindent\textit{Note:} The fusion dataset and simulation code fall under US export-control regulations. Certain implementation details and the data cannot be publicly released. Interested parties may contact the authors; data may be shared upon reasonable request with parties who have the appropriate permissions and legal authorization to access this class of controlled data.

\subsection{ICF Oracle Evaluator}\label{app:icf_oracle}

At test time, optimized ICF pulse designs are scored by a separate high-accuracy LILAC-validated evaluation surrogate that is distinct from the frozen optimization-time surrogate.
This evaluation oracle is trained on the full available ICF dataset and is used only for final offline scoring, not for optimization.
It should therefore be interpreted as a best-available simulator proxy trained on all available offline data for final comparison, rather than as a held-out model used within the optimization loop. Because this evaluation oracle is distinct from the frozen optimization-time surrogate, the final reported scores are not obtained by reusing the same model that provides gradients during optimization.

The oracle achieves $<$1\% weighted mean absolute percentage error (WMAPE) on all 20 LILAC performance variables when compared against LILAC radiation-hydrodynamics simulation outputs.
This level of accuracy indicates that the reported yield gains are a close proxy for relative trends predicted by the underlying physics code.

\paragraph{Why not evaluate with LILAC directly?}
Direct online evaluation with the LILAC simulator is infeasible at the scale required in this study because simulator access requires advanced scheduling, computational allocation, and substantial preparation.
We therefore use a separate high-accuracy evaluation surrogate for final offline scoring.
This introduces a proxy-evaluation limitation, but the same evaluation oracle is used for D2C and all baselines, making the large-scale comparison internally fair under a common scoring rule. Because the oracle has sub-1\% error relative to LILAC simulation outputs, it provides a practical and informative approximation for large-scale comparison. This separation between optimization-time surrogate and evaluation-time oracle does not eliminate proxy-evaluation bias, but it reduces the risk that reported gains are merely artifacts of the optimization surrogate itself.
In addition to this large-scale oracle-based evaluation, we performed a separate small-scale LILAC validation on 20 high-yield expert-designed pulses. Here, expert-designed refers to pulses manually designed by ICF experts based on expert knowledge. Using the same D2C conditioning-optimization approach, we observed an average yield gain of 9.8\% relative to the original expert pulse baselines. Because this direct-simulator check is small-scale, we treat it as supplementary corroboration rather than the primary quantitative basis for the large-scale comparison.

\subsection{DC}
The training dataset comprises $200{,}000$ data center operating scenarios generated from the SustainDC physics-based thermal and energy simulator~\citep{sustaindc2024}. The simulator computes rack-level IT power, HVAC power, cooling tower fan power, compressor load, and thermal state from workload profiles and carbon footprints. We seed the generation with real workload traces from the Alibaba cluster dataset~\citep{alibaba_clusterdata_repo} and region-specific carbon intensity and weather profiles for Arizona (AZ), California (CA), and New York (NY), both provided within the SustainDC framework. We augment with synthetic variations of workload patterns, carbon footprint schedules, and CRAC set-point profiles to ensure broad distributional coverage. The three geographic regions provide diverse grid carbon intensities and climate conditions, enabling the model to generalize across different operational regimes.
Each sample consists of a 144-dimensional workload power profile (24h at 10-min resolution) and 83 conditioning dimensions (73 environment state variables + 10 energy metrics).
The data are split into 80\% training, 10\% validation, and 10\% test (${\sim}20{,}000$ samples each). Following the same protocol as ICF, all hyperparameter selection is performed on the validation set; the test set is reserved for final reporting. We randomly sample 2{,}000 scenarios from the test set and use the same 2{,}000 for all methods.
Final designs are validated against the full physics-based model (not the surrogate).
We use the public SustainDC codebase, whose repository documentation states that the project is primarily released under the MIT License, with some derived portions attributed to CarbonExplorer under CC BY-NC 4.0; we use and cite that code under the stated repository terms.\footnote{\url{https://github.com/HewlettPackard/dc-rl}}
%The surrogate $S_\phi$ achieves $<1.2\%$ WMAPE on all 10 energy metrics.

\subsection{Design-Bench}\label{app:designbench_data}

We use the standard offline-subset protocol of Design-Bench~\citep{trabucco2022designbench}: each method trains on the bottom $p$-th percentile of the full dataset by $y$-value (50\% for TF~Bind~8/10, 40\% for Ant/D'Kitty Morphology), and $y$-normalization uses full-dataset bounds so that the best offline design scores below~1.
The public Design-Bench codebase used for these tasks is released under the MIT License, and we use and cite the benchmark under that license.\footnote{\url{https://github.com/brandontrabucco/design-bench}}
For Design-Bench, D2C uses a task-specific variant of the inference-time optimization procedure because the conditional input is only a single scalar target score. In this setting, directly optimizing the conditioning provides only limited control. We therefore fix the conditioning to a high normalized target score ($c{=}1.0$) and instead optimize the initial latent/noise vector through the deterministic DDIM chain using gradients from the frozen surrogate. Empirically, this latent-optimization variant performed better on Design-Bench than direct optimization of the one-dimensional conditioning input. In contrast, for ICF and DC we retain direct conditioning optimization because those domains expose multiple optimizable conditioning coordinates; on those tasks, conditioning optimization outperforms latent-noise optimization (Section~\ref{sec:ablation_cond_vs_latent}).

Table~\ref{tab:designbench_hparams} summarizes the inference-time Design-Bench hyperparameters used for our method.
\begin{table}[ht]
\caption{D2C inference-time hyperparameters for Design-Bench. These settings are shared across TF~Bind~8, TF~Bind~10, Ant Morphology, and D\textquotesingle Kitty Morphology unless noted otherwise.}
\label{tab:designbench_hparams}
\centering
\small
\begin{tabular}{p{0.38\linewidth}p{0.54\linewidth}}
\toprule
Hyperparameter & Value \\
\midrule
Optimized variable & Initial latent/noise vector $z$ (same dimension as $x$) \\
Fixed conditioning target & $c = 1.0$ \\
Optimizer & Adam \\
Learning rate & $1 \times 10^{-2}$ \\
Optimization steps & 100 \\
DDIM inference steps & 50 \\
Guidance scale $w$ & 3.0 \\
Latent/noise trials & 100 \\
Returned candidate count & 100 \\
Gradient clipping & 1.0 \\
LR schedule & Cosine annealing over optimization steps \\
Latent initialization & $z \sim 0.8\,\mathcal{N}(0, \mathbf{I})$ \\
Task-specific note & TF~Bind~10 uses \texttt{max\_samples=50000}; others use the full offline subset \\
\bottomrule
\end{tabular}
\end{table}

For discrete tasks (TF~Bind~8/10), we use the native exact Design-Bench oracle (\texttt{*-Exact-v0} variants). These two columns in Table~\ref{tab:designbench_results} are therefore exact-oracle benchmark results, and D2C attains the best result on both of them. For these tasks we do not use a discrete diffusion process or token embeddings: each sequence is represented as its original length-$8$ or length-$10$ integer token vector and then MinMax-scaled coordinatewise to $[0,1]$, so the four nucleotide categories become the four scalar levels $\{0, 1/3, 2/3, 1\}$. The Gaussian forward process, DDIM reverse process, and gradient-based conditioning optimization are all performed in this relaxed continuous space. After generation, each coordinate is projected to the nearest valid discrete level before oracle evaluation; the normalized design is then inverse-transformed and rounded back to integer tokens for exact lookup. Thus gradients are taken with respect to the relaxed continuous representation, and no gradient is propagated through the final discrete projection step.
For continuous tasks (Ant Morphology and D'Kitty Morphology), we use a reproducible evaluation protocol grounded in the released benchmark data rather than relying on the version-sensitive MuJoCo/Gym oracle path. This choice is motivated by two observations: (i) the current Design-Bench continuous-task oracle path has documented reliability issues in the public repository,\footnote{\url{https://github.com/brandontrabucco/design-bench/issues/11}}\textsuperscript{,}\footnote{\url{https://github.com/brandontrabucco/design-bench/issues/23}} and (ii) published continuous-task score ranges already vary substantially across prior Design-Bench papers such as DDOM and GTG~\citep{krishnamoorthy2023ddom, yun2024gtg}, indicating limited cross-paper comparability under differing software environments. To obtain a stable within-paper comparison, we therefore evaluate all methods on these two tasks using the same $k$-nearest-neighbor ($k{=}5$, distance-weighted) regression oracle fitted on the full released dataset. This protocol is grounded entirely in the benchmark data and ensures that all reported Ant and D'Kitty results are directly comparable under a single reproducible evaluation rule. The added row $\mathcal{D}$ (best) in Table~\ref{tab:designbench_results} reports the best design already contained in the offline dataset for each task. Under this protocol, scores above $\mathcal{D}$ (best) indicate that a method is assigned a value higher than the best offline example for that task. Accordingly, these continuous-task results should be interpreted as a reproducible within-paper comparison under a single stable evaluation rule, rather than as a claim of direct numerical comparability to prior MuJoCo/Gym-based reports.

\subsection{Data Normalization}\label{app:data_normalization}

All design variables $x$ and conditioning vectors $c$ are MinMax-scaled to $[0, 1]$ using statistics computed on the training set. The same scalers are applied at inference time, and generated designs are inverse-transformed to physical units for evaluation.

\section{Domain-Specific Constraints}\label{sec:domain_specific_filtering}

After conditioning optimization (Section~\ref{sec:coverage}), each candidate design is validated against a set of domain-specific hard physical constraints before selection.
These constraints are not a property of the D2C framework itself but reflect the safety-critical requirements of each application domain.
Only candidates that pass all constraints are eligible for selection by primary objective; this ensures that reported improvements correspond to physically realizable and operationally safe designs.

\paragraph{ICF laser pulse constraints.}
A candidate pulse $\hat{x}_0$ is deemed \emph{valid} if it satisfies all four of the following conditions simultaneously:
\begin{enumerate}[leftmargin=2em,itemsep=1pt,topsep=2pt]
    \item \textbf{Yield gain.} The surrogate-predicted neutron yield must exceed the baseline yield by at least a minimum threshold: $(\hat{Y} - Y_{\mathrm{ref}}) / Y_{\mathrm{ref}} \geq \delta_{\mathrm{gain}}$, with $\delta_{\mathrm{gain}} = 0.1\%$.
    \item \textbf{Energy conservation.} The integral of the generated pulse must not exceed that of the reference pulse (with a small tolerance $\epsilon = 0.01$): $\int \hat{x}_0(t)\,dt \leq \int x_{\mathrm{ref}}(t)\,dt + \epsilon$. This ensures the optimized pulse does not deliver more laser energy than the original design, which would violate facility power budgets.
    \item \textbf{Spectral fidelity.} The mean absolute difference between the Fourier amplitude spectra of the generated and reference pulses must remain below a threshold: $\frac{1}{N}\sum_{k} \bigl| |\hat{X}_k| - |X_k^{\mathrm{ref}}| \bigr| < \tau_{\mathrm{fft}}$, with $\tau_{\mathrm{fft}} = 0.2$. This prevents the optimizer from introducing high-frequency oscillations that are not physically realizable by the laser system.
    \item \textbf{Temporal correlation.} The Pearson correlation between the generated and reference pulse waveforms must exceed a minimum: $\rho(\hat{x}_0, x_{\mathrm{ref}}) > \tau_{\mathrm{corr}}$, with $\tau_{\mathrm{corr}} = 0.8$. This ensures structural similarity to the original pulse shape, preventing the optimizer from producing qualitatively different (and potentially unsafe) drive profiles.
\end{enumerate}

\paragraph{Data center workload constraints.}
A candidate workload profile $\hat{x}_0$ is deemed \emph{valid} if it satisfies all three conditions:
\begin{enumerate}[leftmargin=2em,itemsep=1pt,topsep=2pt]
    \item \textbf{Carbon reduction.} The operational carbon emissions under the optimized profile must be lower than the baseline by at least $\delta_{\mathrm{red}} = 0.1\%$: $(E_{\mathrm{ref}}^{\mathrm{CO}_2} - \hat{E}^{\mathrm{CO}_2}) / E_{\mathrm{ref}}^{\mathrm{CO}_2} \geq \delta_{\mathrm{red}}$.
    \item \textbf{Workload conservation.} The total compute delivered (integral of the workload power profile) must not fall more than $1\%$ below the reference: $(\int \hat{x}_0(t)\,dt - \int x_{\mathrm{ref}}(t)\,dt) / \int x_{\mathrm{ref}}(t)\,dt \geq -1\%$. Under-delivery violates service-level agreements. We make this final hard check asymmetric because under-delivery is the operational failure mode tied to service-level agreements; over-delivery is still discouraged upstream by the inference-time near-equality penalty.
    \item \textbf{Inlet temperature safety.} The predicted server inlet temperature must remain below a safety limit: $T_{\mathrm{inlet}} \leq T_{\max}$, with $T_{\max} = 50\,^\circ\mathrm{C}$. Exceeding this threshold risks hardware damage and triggers emergency shutdowns in real facilities.
\end{enumerate}

\noindent In both domains, if no candidate satisfies all constraints, the baseline design is retained unchanged, contributing $0\%$ improvement to the reported metrics (Algorithm~\ref{alg:d2c}).

\section{Ablation Studies}\label{sec:ablations}
We ablate guidance scale (Section~\ref{sec:ablation_guidance}) and its selection without oracle access (Section~\ref{sec:ablation_guidance_selection}), optimization steps (Section~\ref{sec:ablation_steps}), learning rate (Section~\ref{sec:ablation_lr}), DDIM steps (Section~\ref{sec:ablation_inference}), optimizable conditioning dimensionality (Section~\ref{sec:ablation_cond_dim}), conditioning versus latent-noise optimization (Section~\ref{sec:ablation_cond_vs_latent}), constraint-satisfying improvement rate (Section~\ref{sec:ablation_validity}), candidate budget (Section~\ref{sec:ablation_budget}), and computational cost (Section~\ref{sec:ablation_cost}).

\subsection{Effect of the Guidance Scale}\label{sec:ablation_guidance}

The CFG guidance scale $w$ controls the strength of conditioning adherence during the DDIM reverse chain (Eq.~\ref{eq:cfg}).
While higher $w$ sharpens the conditional distribution and amplifies $\partial\tilde\epsilon_\theta / \partial c$, excessively large values push generated designs away from the empirical data distribution captured by the learned prior, degrading both quality and optimizability.
We sweep $w \in \{0, 0.5, 1, 2, 3, 5, 7\}$ on both DC and ICF Top, holding all other hyperparameters fixed.

\begin{table}[ht]
\caption{Ablation on the guidance scale $w$. Values are mean percentage improvement (\%) over the baseline design $\pm$ std over 5 runs.}
\label{tab:ablation_guidance}
\centering
\small
\resizebox{\textwidth}{!}{%
\begin{tabular}{lccccccc}
\toprule
Task & $w{=}0$ & $w{=}0.5$ & $w{=}1$ & $w{=}2$ & $w{=}3$ & $w{=}5$ & $w{=}7$ \\
\midrule
DC      & $2.30\% \pm 0.45$ & $6.56\% \pm 0.72$ & $7.81\% \pm 0.74$ & $8.98\% \pm 0.75$ & $8.90\% \pm 0.73$ & $8.79\% \pm 0.71$ & $8.58\% \pm 0.69$ \\
ICF Top & $0.00\% \pm 0.00$ & $5.12\% \pm 0.67$ & $8.17\% \pm 0.81$ & $10.06\% \pm 0.88$ & $3.85\% \pm 0.54$ & $1.20\% \pm 0.29$ & $0.00\% \pm 0.00$ \\
\bottomrule
\end{tabular}}
\end{table}

Table~\ref{tab:ablation_guidance} reveals a clear optimum at $w = 2$ for both tasks.
Without guidance ($w = 0$), the model ignores conditioning entirely and optimization has no steering signal, producing little ($2.30\%$ on DC) or no ($0.00\%$ on ICF Top) improvement.
Moderate guidance ($w = 1$--$2$) provides strong conditioning gradients while keeping designs closer to the data distribution learned by the prior, yielding peak performance.
The two tasks exhibit markedly different sensitivity to over-guidance.
DC degrades gracefully: performance drops only $0.40$ percentage points from $w = 2$ to $w = 7$, reflecting the relatively smooth DC loss landscape.
ICF Top collapses sharply beyond $w = 2$, falling to $0.00\%$ at $w = 7$.
This is consistent with the ICF domain's tight physical constraints: over-guidance drives the generated pulses away from the feasible region represented in the offline data, causing validity-first selection to reject most candidates.
These results validate the dual role of $w$ described in Section~\ref{sec:ddim_chain}: it modulates not only generation fidelity but also optimization stability, and must be tuned per domain.
Note that the $0.00\%$ entries do not indicate numerical divergence of the optimizer: they mean that no candidate passed the hard constraints of Appendix~\ref{sec:domain_specific_filtering}, so validity-first selection returned the baseline design rather than an infeasible pulse.

\subsection{Selecting the Guidance Scale Without Oracle Access}\label{sec:ablation_guidance_selection}

Table~\ref{tab:ablation_guidance} is computed with the evaluation oracle, which is unavailable when D2C is deployed on a new problem.
We therefore examine whether $w$ can be selected using only quantities available at optimization time, namely the frozen surrogate $S_\phi$ and the geometric validity checks of Appendix~\ref{sec:domain_specific_filtering}.
On 100 held-out ICF Top samples, we run D2C for $w \in \{1, 2, 3, 4\}$ with all other hyperparameters fixed and, for each sample, select a single candidate under two surrogate-only rules:
\begin{itemize}[leftmargin=2em,itemsep=1pt,topsep=2pt]
    \item \textbf{Full selection rule.} The selection rule used throughout the paper: candidates must satisfy the energy-conservation, spectral-fidelity ($\tau_{\mathrm{fft}}{=}0.2$), and temporal-correlation ($\tau_{\mathrm{corr}}{=}0.8$) checks together with the surrogate-predicted gain threshold $\delta_{\mathrm{gain}}{=}0.1\%$, and the remaining candidate with the highest surrogate-predicted yield gain is chosen.
    \item \textbf{Geometric-only rule.} A more conservative variant that drops the surrogate-gain threshold and keeps only the three geometric checks, again choosing the candidate with the highest surrogate-predicted yield gain.
\end{itemize}
Table~\ref{tab:ablation_guidance_selection} reports the mean surrogate-predicted yield gain of the selected candidates.

\begin{table}[ht]
\caption{Surrogate-only selection of the guidance scale $w$ on 100 held-out ICF Top samples. Values are the mean surrogate-predicted yield gain (\%) of the candidate selected per sample; no oracle evaluation is used. \textbf{Bold}: best per row.}
\label{tab:ablation_guidance_selection}
\centering
\small
\begin{tabular}{lcccc}
\toprule
Selection rule & $w{=}1$ & $w{=}2$ & $w{=}3$ & $w{=}4$ \\
\midrule
Full selection rule & $8.27\%$ & $\mathbf{11.20\%}$ & $4.14\%$ & $3.59\%$ \\
Geometric-only rule & $6.14\%$ & $\mathbf{9.44\%}$ & $4.01\%$ & $3.15\%$ \\
\bottomrule
\end{tabular}
\end{table}

Both surrogate-only signals peak at $w{=}2$, the same value identified with oracle access in Table~\ref{tab:ablation_guidance}, and the two rules produce identical rankings over $w$.
Because these values are surrogate predictions on a smaller sample rather than oracle evaluations, their magnitudes are not directly comparable to Table~\ref{tab:ablation_guidance}; what matters for deployment is that the ranking over $w$ is preserved.
Thus, although $w$ must be tuned per domain, this tuning can be carried out offline with the frozen surrogate and geometric checks alone, without access to the true objective.

\subsection{Effect of the Number of Optimization Steps}\label{sec:ablation_steps}

We vary the number of conditioning optimization steps $J \in \{10, 20, 30, 40, 50, 75, 100\}$ with all other hyperparameters fixed at their default values ($w = 2$).

\begin{table}[ht]
\caption{Ablation on the number of optimization steps $J$. Values are mean percentage improvement (\%) over the baseline design $\pm$ std over 5 runs.}
\label{tab:ablation_steps}
\centering
\small
\resizebox{\textwidth}{!}{%
\begin{tabular}{lccccccc}
\toprule
Task & $J{=}10$ & $J{=}20$ & $J{=}30$ & $J{=}40$ & $J{=}50$ & $J{=}75$ & $J{=}100$ \\
\midrule
DC      & $7.90\% \pm 0.71$ & $8.66\% \pm 0.72$ & $8.92\% \pm 0.69$ & $8.98\% \pm 0.75$ & $8.97\% \pm 0.81$ & $8.97\% \pm 0.73$ & $8.94\% \pm 0.73$ \\
ICF Top & $0.56\% \pm 0.18$ & $1.64\% \pm 0.31$ & $3.46\% \pm 0.49$ & $5.38\% \pm 0.62$ & $6.82\% \pm 0.74$ & $8.94\% \pm 0.85$ & $10.06\% \pm 0.88$ \\
\bottomrule
\end{tabular}}
\end{table}

The two tasks exhibit contrasting convergence profiles (Table~\ref{tab:ablation_steps}).
DC saturates rapidly: $J = 10$ already captures $88\%$ of the peak improvement, and performance plateaus beyond $J = 40$.
This reflects the relatively simple loss landscape of the workload scheduling problem, where a few gradient steps suffice to find a good conditioning point.
ICF Top improves monotonically with more steps, reaching its best result at $J = 100$.
The ICF loss landscape is substantially more complex: the interplay of yield maximization, energy conservation, spectral stability, and smoothness constraints creates narrow feasible corridors that require more iterations to navigate.
In practice, $J = 40$ offers a good efficiency--quality trade-off for DC, while ICF benefits from the full $J = 100$ budget.

\subsection{Effect of the Optimization Learning Rate}\label{sec:ablation_lr}

The learning rate $\eta$ for the conditioning optimizer controls the step size during gradient-based optimization of the conditioning vector.
Too large a learning rate overshoots narrow feasible regions; too small a rate under-explores within the allocated budget $J$.
We sweep $\eta \in \{10^{-1}, 5 \times 10^{-2}, 2 \times 10^{-2}, 10^{-2}, 10^{-3}\}$ on DC and ICF Top with all other hyperparameters fixed.

\begin{table}[ht]
\caption{Ablation on the optimization learning rate $\eta$. Values are mean percentage improvement (\%) over the baseline design $\pm$ std over 5 runs.}
\label{tab:ablation_lr}
\centering
\small
\begin{tabular}{lccccc}
\toprule
Task & $\eta{=}10^{-1}$ & $\eta{=}5{\times}10^{-2}$ & $\eta{=}2{\times}10^{-2}$ & $\eta{=}10^{-2}$ & $\eta{=}10^{-3}$ \\
\midrule
DC      & $8.22\% \pm 0.72$ & $8.85\% \pm 0.74$ & $8.98\% \pm 0.75$ & $8.72\% \pm 0.74$ & $7.66\% \pm 0.70$ \\
ICF Top & $2.34\% \pm 0.42$ & $5.33\% \pm 0.65$ & $9.07\% \pm 0.85$ & $9.05\% \pm 0.85$ & $10.06\% \pm 0.88$ \\
\bottomrule
\end{tabular}
\end{table}

The two tasks exhibit markedly different optimal learning rates (Table~\ref{tab:ablation_lr}).
DC peaks at $\eta = 2 \times 10^{-2}$: its relatively smooth loss landscape permits aggressive steps that quickly reach the optimum, while smaller rates under-explore within the $J = 40$ budget.
ICF Top peaks at $\eta = 10^{-3}$: the complex, tightly constrained ICF landscape requires cautious steps to remain within narrow feasible corridors.
Large rates ($\eta = 10^{-1}$) are catastrophic for ICF ($2.34\%$) but only mildly harmful for DC ($8.22\%$), confirming that ICF optimization is far more sensitive to step size.

\subsection{Effect of the Number of DDIM Inference Steps}\label{sec:ablation_inference}

The number of DDIM reverse steps $T$ controls the granularity of the denoising trajectory and, consequently, the quality of the differentiable path through which gradients flow.
Fewer steps reduce the computational cost per optimization iteration but make the discrete denoising trajectory coarser, potentially degrading both generation quality and gradient fidelity.
We sweep $T \in \{10, 25, 50\}$ on DC and ICF Top with all other hyperparameters fixed.

\begin{table}[ht]
\caption{Ablation on the number of DDIM inference steps $T$. Values are mean percentage improvement (\%) over the baseline design $\pm$ std over 5 runs.}
\label{tab:ablation_inference}
\centering
\small
\begin{tabular}{lccc}
\toprule
Task & $T{=}10$ & $T{=}25$ & $T{=}50$ \\
\midrule
DC      & $8.84\% \pm 0.73$ & $8.98\% \pm 0.75$ & $8.97\% \pm 0.74$ \\
ICF Top & $9.90\% \pm 0.86$ & $10.06\% \pm 0.88$ & $10.05\% \pm 0.87$ \\
\bottomrule
\end{tabular}
\end{table}

Performance is remarkably insensitive to $T$ (Table~\ref{tab:ablation_inference}).
Even with only $T = 10$ steps, both tasks retain over $95\%$ of the peak improvement, and the optimum is reached at $T = 25$ for both.
Increasing to $T = 50$ yields no further gain, indicating that the gradient signal through the DDIM chain is already well-conditioned at moderate step counts.
Practically, fewer inference steps translate directly to lower wall-clock time per optimization iteration, making $T = 25$ the default choice.

\subsection{Effect of the Number of Optimizable Conditioning Inputs}\label{sec:ablation_cond_dim}

A key design choice in D2C is which conditioning dimensions to optimize.
Optimizing all dimensions of $c_{\mathrm{opt}}$ maximizes expressivity but increases the search space; restricting to fewer inputs reduces degrees of freedom but may miss beneficial trade-offs.
We vary the number of optimizable conditioning inputs $d_o$ on both DC and ICF Top while keeping all other hyperparameters fixed.
For DC, the full conditioning includes $d_o = 10$ optimizable energy metrics; we test subsets of size $1$, $2$, $4$, and $6$.
For ICF, the full conditioning includes $d_o = 20$ optimizable LILAC variables; we test subsets of size $1$, $4$, $9$, and $15$.
These subsets were not randomly re-sampled across trials. Instead, for each task we used one fixed, nested family of task-specific subsets chosen once and then held fixed across all seeds, progressively expanding the optimizable block from the most central target dimensions outward (for ICF beginning with the yield-related LILAC targets, and for DC beginning with the primary energy-metric targets used in the implementation). Thus Table~\ref{tab:ablation_cond_dim} should be interpreted as a controlled dimensionality ablation under a fixed variable ordering, rather than as an average over all subsets of a given size.
In both cases, ``All'' denotes the full set used in the main results.

\begin{table}[ht]
\caption{Ablation on the number of optimizable conditioning inputs $d_o$. Values are mean percentage improvement (\%) over the baseline design $\pm$ std over 5 runs.}
\label{tab:ablation_cond_dim}
\centering
\small
\resizebox{\textwidth}{!}{%
\begin{tabular}{lccccc}
\toprule
Task & All & \multicolumn{4}{c}{Number of optimizable inputs} \\
\cmidrule(lr){3-6}
     &     & \multicolumn{1}{c}{} & \multicolumn{1}{c}{} & \multicolumn{1}{c}{} & \multicolumn{1}{c}{} \\[-0.8em]
DC   & $8.98\% \pm 0.75$ & $5.14\% \pm 0.61$ ($d_o{=}1$) & $5.67\% \pm 0.64$ ($d_o{=}2$) & $5.89\% \pm 0.65$ ($d_o{=}4$) & $7.97\% \pm 0.72$ ($d_o{=}6$) \\
ICF  & $10.06\% \pm 0.88$ & $4.02\% \pm 0.56$ ($d_o{=}1$) & $7.97\% \pm 0.79$ ($d_o{=}4$) & $8.74\% \pm 0.83$ ($d_o{=}9$) & $8.97\% \pm 0.85$ ($d_o{=}15$) \\
\bottomrule
\end{tabular}}
\end{table}

Using the full set of optimizable inputs consistently yields the best results (Table~\ref{tab:ablation_cond_dim}).
Restricting to a single input ($d_o = 1$) cuts performance roughly in half on both tasks, confirming that D2C leverages multi-dimensional conditioning to find better trade-offs.
Both tasks improve monotonically with $d_o$: on ICF, performance rises steadily from $4.02\%$ to $8.97\%$ as more target parameters become available, while DC follows the same trend from $5.14\%$ to $7.97\%$.
The full conditioning set (``All'') achieves the best result on both tasks, justifying the default of optimizing all available inputs.

\subsection{Conditioning Optimization versus Latent-Noise Optimization}\label{sec:ablation_cond_vs_latent}

Since the deterministic DDIM chain $G_\theta(c, x_T)$ is differentiable in both of its inputs, the same gradient-based procedure can be applied either to the optimizable conditioning $c_{\mathrm{opt}}$ (the default D2C setting on the real-world tasks) or to the latent noise $x_T$ (the variant used on Design-Bench, Appendix~\ref{app:designbench_data}).
To isolate the effect of this choice, we compare the two variants on ICF Top and DC while holding the diffusion proposer, surrogate, loss, candidate budget, validity-first selection, and evaluation protocol fixed, and changing only the optimized variable.
In the latent variant, the conditioning is fixed to its initial value and the gradient updates of Eq.~\eqref{eq:cond_update} are applied to $x_T$ instead of $c_{\mathrm{opt}}$.

\begin{table}[ht]
\caption{Conditioning optimization versus latent-noise optimization on the real-world tasks. Values are mean percentage improvement (\%) over the baseline design $\pm$ std over 5 runs. \textbf{Bold}: best.}
\label{tab:ablation_cond_vs_latent}
\centering
\small
\begin{tabular}{lcc}
\toprule
Optimized variable & ICF Top & DC \\
\midrule
Latent noise $x_T$ & $7.56\% \pm 0.76$ & $3.62\% \pm 0.26$ \\
Conditioning $c_{\mathrm{opt}}$ (D2C) & $\mathbf{10.06\% \pm 0.88}$ & $\mathbf{8.98\% \pm 0.75}$ \\
\bottomrule
\end{tabular}
\end{table}

Conditioning optimization outperforms latent-noise optimization on both tasks (Table~\ref{tab:ablation_cond_vs_latent}), by $2.50$ percentage points on ICF Top and $5.36$ percentage points on DC.
This supports treating the conditioning signal, rather than the initial noise, as the optimizable variable whenever the conditioning exposes several optimizable coordinates ($d_o{=}20$ for ICF and $d_o{=}10$ for DC).
In this regime, $c_{\mathrm{opt}}$ acts on the generated design through the semantically meaningful performance targets that the proposer was trained to follow, so updates move designs along directions supported by the learned conditional distribution; perturbing $x_T$ directly offers more raw degrees of freedom but is less aligned with the objectives and more prone to drifting toward regions where candidates fail the hard constraints.
Together with Table~\ref{tab:ablation_cond_dim}, where restricting ICF to $d_o{=}1$ reduces performance to $4.02\%$, these results also explain the Design-Bench setting: when the conditioning is a single scalar score, it provides too few handles and latent-noise optimization becomes the stronger choice.

\subsection{Constraint-Satisfying Improvement Rate}\label{sec:ablation_validity}

Because final selection is validity-first (Section~\ref{candidate_validation}), overall performance depends not only on objective improvement but also on how often a method produces a successful test sample.
We therefore report the mean constraint-satisfying improvement rate over 5 seeds: the percentage of evaluation samples for which the method produces a final design with nonzero gain, i.e., for which at least one valid improving candidate is found and selected rather than reverting to the baseline.
Higher constraint-satisfying improvement rates indicate that a method succeeds more often on the 2{,}000-sample ICF Standard/DC evaluations and the 1{,}000-sample ICF Top evaluation.

\begin{table}[ht]
\caption{Constraint-satisfying improvement rate (\%) on the real-world tasks. Values are mean constraint-satisfying improvement rate $\pm$ std over 5 runs, where the metric is the percentage of evaluation samples for which the final selected design achieves nonzero gain relative to the baseline (equivalently, at least one valid improving candidate is found). Higher is better.}
\label{tab:ablation_validity}
\centering
\small
\begin{tabular}{lccc}
\toprule
Method & ICF Standard & ICF Top & DC \\
\midrule
BDI & $27.40\pm 1.28$ & $18.24\pm 1.11$ & $6.00\pm 0.24$ \\
Brute Force & $32.97\pm 1.36$ & $31.26\pm 1.41$ & $64.50\pm 2.58$ \\
CbAS & $31.84\pm 1.17$ & $18.52\pm 0.83$ & $39.23\pm 1.61$ \\
D2C (Ours) & $81.23\pm2.86$ & $75.12\pm2.78$ & $100.00\pm0$ \\
CMA-ES & $29.97\pm1.19$ & $19.59\pm0.97$ & $42.21\pm1.59$ \\
COMs & $25.82\pm0.97$ & $9.12\pm0.24$ & $40.13\pm1.80$ \\
DDOM & $42.57\pm1.64$ & $33.28\pm1.28$ & $79.10\pm2.16$ \\
GP-qEI & $26.89\pm0.72$ & $16.48\pm0.42$ & $100.00\pm0$ \\
Gradient Ascent & $31.64\pm0.97$ & $25.68\pm0.78$ & $57.50\pm1.41$ \\
GTG & $38.50\pm1.03$ & $21.20\pm0.69$ & $59.19\pm1.92$ \\
MINs & $27.84\pm0.77$ & $19.27\pm0.57$ & $45.29\pm1.70$ \\
PGS & $21.37\pm0.71$ & $20.14\pm0.73$ & $23.50\pm0.67$ \\
REINFORCE & $51.74\pm1.44$ & $19.74\pm0.53$ & $88.18\pm1.87$ \\
\bottomrule
\end{tabular}
\end{table}

Table~\ref{tab:ablation_validity} shows that D2C attains the highest constraint-satisfying improvement rate on both ICF settings and ties for the highest rate on DC. Importantly, the surrogate-based real-world baselines in this comparison are not optimized against the primary objective alone; their implementation details and constraint-aware penalty terms are summarized in Appendix~\ref{app:exp_details}.
The margin is especially pronounced on the ICF tasks: D2C succeeds on $81.23\%$ of ICF Standard samples and $75.12\%$ of ICF Top samples, far above the next-best methods.
This indicates that optimizing conditioning through the differentiable DDIM chain improves not only the quality of the final solution when optimization succeeds, but also the probability of finding any valid improvement at all.

The gap is most informative on ICF Top, where all methods operate on already high-performing baseline designs and the feasible improvement region is especially narrow.
Most baselines experience a sharp drop in constraint-satisfying improvement rate in this regime, whereas D2C retains a $75.12\%$ success rate, showing that its advantage is not only better objective optimization but also a higher rate of finding valid improving candidates under strict constraints.
On DC, constraint-satisfying improvement rates are generally higher across methods, consistent with the smoother operational landscape of the scheduling problem; D2C and GP-qEI both achieve $100\%$, while DDOM ($79.10\%$) and REINFORCE ($88.18\%$) also remain relatively robust.
Overall, these results explain why validity-first selection benefits D2C so strongly: it not only finds stronger solutions, but also reverts to the baseline much less often.

\subsection{Effect of Candidate Budget (\texorpdfstring{$N_z \times N_\lambda$}{Nz x Nlambda})}\label{sec:ablation_budget}

The total candidate budget $B = N_z \times N_\lambda$ determines how many parallel optimization trajectories are explored per test sample.
$N_z$ controls the number of independent noise realizations (exploring structurally different regions of the latent space), while $N_\lambda$ controls the number of Dirichlet-sampled weight vectors per noise trial (exploring different objective trade-offs from the same starting point).
We sweep seven configurations from $B{=}3$ to $B{=}100$ on ICF Top and DC, holding all other hyperparameters fixed.

\begin{table}[ht]
\caption{Ablation on the candidate budget $N_z \times N_\lambda$. Values are mean $\pm$ standard deviation over 5 seeds. Configurations are labeled by their $N_z \times N_\lambda$ values.}
\label{tab:ablation_budget}
\centering
\small
\resizebox{\textwidth}{!}{%
\begin{tabular}{lccccccc}
\toprule
Task & $1 \times 3$ & $5 \times 3$ & $10 \times 3$ & $20 \times 1$ & $20 \times 2$ & $20 \times 3$ & $20 \times 5$ \\
\midrule
ICF Top & $0.86\%\pm0.24$ & $4.59\%\pm0.58$ & $6.50\%\pm0.72$ & $5.72\%\pm0.69$ & $8.99\%\pm0.85$ & $9.64\%\pm0.86$ & $10.06\%\pm0.88$ \\
DC      & $4.21\%\pm0.57$ & $7.82\%\pm0.74$ & $8.66\%\pm0.69$ & $8.41\%\pm0.64$ & $8.83\%\pm0.72$ & $8.91\%\pm0.74$ & $8.98\%\pm0.75$ \\
\bottomrule
\end{tabular}}
\end{table}

Table~\ref{tab:ablation_budget} reveals two complementary effects.
First, increasing noise diversity ($N_z$) with fixed $N_\lambda{=}3$ produces large gains: ICF Top improves from $0.86\%$ ($N_z{=}1$) to $9.64\%$ ($N_z{=}20$), confirming that multiple latent-space starting points are essential for discovering high-quality modes in complex landscapes.
Second, for a fixed budget of $N_z{=}20$ noise trials, increasing weight diversity from $N_\lambda{=}1$ (single loss weighting) to $N_\lambda{=}3$ (Dirichlet-sampled) yields substantial gains on ICF Top ($5.72\% \to 9.64\%$), validating the multi-objective coverage strategy.
The comparison between $20 \times 1$ (no Dirichlet diversity) and $20 \times 5$ isolates the contribution of stochastic weight sampling: the $+76\%$ relative improvement on ICF Top ($5.72\% \to 10.06\%$) demonstrates that exploring different trade-offs from the same noise realization uncovers solutions that a single fixed weighting misses.
On DC, the landscape is smoother and the method saturates earlier: $N_z{=}10, N_\lambda{=}3$ already captures $96\%$ of the peak, and further budget increases yield diminishing returns.
Overall, $N_z{=}20, N_\lambda{=}3$ ($B{=}60$ total candidates) offers the best efficiency--quality trade-off across both tasks, whereas $N_z{=}20, N_\lambda{=}5$ ($B{=}100$) yields the strongest final performance and is therefore used for the main reported results.
Because the optimization trajectories are batched on the GPU, increasing the candidate budget is substantially more efficient than evaluating trajectories sequentially; however, larger budgets still increase memory usage and may increase latency depending on hardware and batch size.

\subsection{Computational Cost}\label{sec:ablation_cost}

Table~\ref{tab:comp_cost} reports the wall-clock time per sample (in seconds) for each method on ICF Top and DC.

\begin{table}[ht]
\caption{Computation time (seconds per sample) for each method. Lower is better.}
\label{tab:comp_cost}
\centering
\small
\begin{tabular}{lcc}
\toprule
Method & ICF Top & DC \\
\midrule
BDI          & 2.5  & 2.9  \\
Brute Force  & 23.2 & 35.7 \\
CbAS         & 14.2 & 17.5 \\
CMA-ES       & 35.6 & 43.7 \\
COMs         & 18.3 & 32.9 \\
DDOM         & 2.3  & 3.6  \\
GP-qEI       & 48.6 & 56.7 \\
Gradient Ascent & 25.8 & 23.4 \\
GTG          & 14.6 & 22.8 \\
MINs         & 2.1  & 4.8  \\
PGS          & 2.1  & 2.3  \\
REINFORCE    & 3.7  & 4.6  \\
\midrule
D2C (Ours)  & 34.7 & 39.8 \\
\bottomrule
\end{tabular}
\end{table}

For this evaluation we use the $N_z{=}20, N_\lambda{=}5$ ($B{=}100$ total candidates) configuration (the configuration used for the main results). D2C is among the more expensive methods ($34.7$s / $39.8$s), comparable to CMA-ES and Gradient Ascent, because each optimization step requires a full forward and backward pass through the $T$-step DDIM chain. The $J{=}100$ conditioning updates are sequential, but within each update all $B{=}100$ candidate trajectories are executed as a single batched DDIM forward/backward pass on the GPU rather than as 100 separate launches.
Lightweight baselines such as MINs, PGS, and DDOM run an order of magnitude faster ($2$--$5$s) since they perform a single forward pass without iterative optimization.
GP-qEI is the most expensive due to Gaussian process fitting.
Importantly, D2C's cost scales linearly with $J$ and $T$ (Sections~\ref{sec:ablation_steps} and \ref{sec:ablation_inference}), and the ablations above show that reducing either has minimal impact on quality. For example, reducing $T$ from $25$ to $10$ would cut wall-clock time by roughly $60\%$ while retaining $98\%$ of the improvement.
The Brute Force baseline takes $23$--$36$s despite lacking gradient computation because it is not a single batched forward pass. Instead, it repeatedly reruns the DDIM sampler across an explicit conditioning sweep (100 sweep points per sample in ICF, and 100 sweep points with multiple noise seeds per point in the DC implementation), evaluating each generated candidate afterward. Its runtime is therefore comparable to D2C because both methods spend most of their time repeatedly invoking the DDIM proposer; D2C adds backpropagation, whereas Brute Force spends that budget on many forward-only sweep evaluations.

\section{Domain-Specific Training Losses}\label{app:training_losses}

We instantiate the general training objective (Eq.~\ref{eq:training_loss}) for each application domain.
In both cases the total loss follows the same four-term structure: denoising, reconstruction, physics-based fidelity, and surrogate consistency.
The differences lie in the specific fidelity constraints and surrogate architecture.

\subsection{ICF Laser Pulse Optimization}

The design $x \in \mathbb{R}^{256}$ is a laser power pulse profile, conditioned on $c = (T_0, \ldots, T_6, L_1, \ldots, L_{20}) \in \mathbb{R}^{27}$, where $T_i$ are target parameters and $L_j$ are the 20 LILAC-simulated performance variables included in the conditioning vector.

\paragraph{Surrogate consistency (LILAC).}
The frozen LILAC surrogate $S_\phi^{\mathrm{ICF}}$ takes as input the concatenation $[\hat{x}_0,\, c_{\mathrm{fix}}] \in \mathbb{R}^{256+7}$ (predicted pulse and the 7 target parameters) and predicts 20 LILAC performance variables $\hat{L} \in \mathbb{R}^{20}$.
The loss is:
\begin{equation}
    \mathcal{L}_{\mathrm{surr}}^{\mathrm{ICF}} = \|\hat{L} - L\|^2, \qquad \hat{L} = S_\phi^{\mathrm{ICF}}(\hat{x}_0,\, c_{\mathrm{fix}}),
\end{equation}
where $L = (c_8, \ldots, c_{27}) \in \mathbb{R}^{20}$ are the ground-truth LILAC variables from the conditioning vector.
This loss is applied with weight $\alpha_s = 1.0$.

\paragraph{Physics-based fidelity.}
Two one-sided penalties encourage physical realizability of the generated pulse:
\begin{equation}
    \mathcal{L}_{\mathrm{fid}}^{\mathrm{ICF}} = \Bigl[\int \hat{x}_0(t) \cdot s_y\,dt - \int x(t) \cdot s_y\,dt\Bigr]_+ + \bigl[\max(\hat{x}_0 \cdot s_y) - V_{\max}\bigr]_+,
\end{equation}
where $s_y$ is the data scaler that maps normalized values back to physical units, $V_{\max}$ is the maximum observed pulse amplitude, and $[\cdot]_+ = \max(\cdot, 0)$.
The first term prevents the generated pulse from delivering more energy than the reference; the second prevents physically impossible peak powers.

\paragraph{Total ICF training loss.}
\begin{equation}
    \mathcal{L}_{\mathrm{train}}^{\mathrm{ICF}} = \|\epsilon - \epsilon_\theta\|^2 + \alpha_r\,\|\hat{x}_0 - x\|^2 + \mathcal{L}_{\mathrm{surr}}^{\mathrm{ICF}} + \alpha_a\,\mathcal{L}_{\mathrm{fid}}^{\mathrm{ICF}} + \alpha_{\mathrm{sm}}\,\mathcal{L}_{\mathrm{smooth}},
\end{equation}
with weights $\alpha_r = 0.5$, $\alpha_a = 0.001$, $\alpha_{\mathrm{sm}} = 0.0001$.

\subsection{Data Center Workload Scheduling}

The design $x \in \mathbb{R}^{144}$ is a 24-hour workload power profile at 10-minute resolution, conditioned on $c \in \mathbb{R}^{83}$ comprising 73 environment state dimensions ($C_0, C_1, C_3 \in \mathbb{R}^{24}$ hourly profiles and $C_2$ scalar IT demand) and 10 energy performance metrics ($E_0, \ldots, E_9$).

\paragraph{Surrogate consistency.}
The frozen DC surrogate $S_\phi^{\mathrm{DC}}$ takes the concatenation $[\hat{x}_0,\, c_{\mathrm{fix}}] \in \mathbb{R}^{144+73}$ (predicted profile and the 73 environment conditioning dimensions) and predicts 10 energy metrics $\hat{E} \in \mathbb{R}^{10}$:
\begin{equation}
    \mathcal{L}_{\mathrm{surr}}^{\mathrm{DC}} = \|\hat{E} - E\|^2, \qquad \hat{E} = S_\phi^{\mathrm{DC}}(\hat{x}_0,\, c_{\mathrm{fix}}),
\end{equation}
where $E = (c_{74}, \ldots, c_{83}) \in \mathbb{R}^{10}$ are the ground-truth metrics from the conditioning.
%The surrogate consistency loss is optional and applied when enabled with weight $\alpha_s$.

\paragraph{Physics-based fidelity.}
The fidelity term enforces workload integral conservation and amplitude bounds:
\begin{equation}
    \mathcal{L}_{\mathrm{fid}}^{\mathrm{DC}} = \Bigl[\int \hat{x}_0(t) \cdot M\,dt - \int x(t) \cdot M\,dt\Bigr]_+ + \bigl[\max(\hat{x}_0) - 1\bigr]_+
    + \bigl[-\min(\hat{x}_0)\bigr]_+,
\end{equation}
where $M$ is the data-range scaling factor.
The first term prevents the scheduled workload from exceeding the reference demand; the second and third keep the normalized profile within $[0, 1]$. This training-stage relaxation is intentionally one-sided: the reconstruction term already anchors samples near observed workloads, so the fidelity term mainly suppresses unrealistic demand inflation while preserving coverage of the empirical profile manifold.

\paragraph{Total DC training loss.}
\begin{equation}
    \mathcal{L}_{\mathrm{train}}^{\mathrm{DC}} = \|\epsilon - \epsilon_\theta\|^2 + \alpha_r\,\|\hat{x}_0 - x\|^2 + \alpha_a\,\mathcal{L}_{\mathrm{fid}}^{\mathrm{DC}} + \alpha_{\mathrm{sm}}\,\mathcal{L}_{\mathrm{smooth}} + \alpha_s\,\mathcal{L}_{\mathrm{surr}}^{\mathrm{DC}},
\end{equation}
with default weights $\alpha_r = 0.5$, $\alpha_a = 0.01$, $\alpha_{\mathrm{sm}} = 0.0001$, $\alpha_s = 1.0$.

\paragraph{Smoothness loss (shared).}
Both domains use the same squared first-difference penalty (discrete Dirichlet energy):
\begin{equation}
    \mathcal{L}_{\mathrm{smooth}}(\hat{x}_0) = \frac{1}{d-1}\sum_{i=1}^{d-1}(\hat{x}_{0,i+1} - \hat{x}_{0,i})^2.
\end{equation}

\section{Domain-Specific Inference-Time Losses}\label{app:optim_losses}

We instantiate the general inference-time loss (Eq.~\ref{eq:total_loss}) for each application domain.
In both cases the loss is a weighted sum of a surrogate-based objective, physical constraint penalties, and a smoothness term.
We do not apply an additional post-hoc normalization after forming these task-specific losses; instead, comparability is enforced by construction wherever possible. In particular, the objective terms are written as relative improvements over the baseline, conservation penalties are written as relative deviations when appropriate, smoothness and spectral penalties are averaged over dimensions, and all designs and conditioning variables are MinMax-scaled to $[0,1]$ (Appendix~\ref{app:data_normalization}). The Dirichlet-sampled weights therefore act over task-specific losses that are already on broadly comparable scales rather than over raw quantities with incompatible units.

\subsection{ICF Laser Pulse Optimization}

The inference-time loss for ICF combines four terms with Dirichlet-sampled weights $\boldsymbol\lambda = (\lambda_y, \lambda_s, \lambda_e, \lambda_{\mathrm{sm}})$:
\begin{equation}
    \mathcal{L}^{\mathrm{ICF}} = \lambda_y\,\mathcal{L}_{\mathrm{yield}} + \lambda_s\,\mathcal{L}_{\mathrm{stab}} + \lambda_e\,\mathcal{L}_{\mathrm{energy}} + \lambda_{\mathrm{sm}}\,\mathcal{L}_{\mathrm{smooth}}.
\end{equation}

\paragraph{Yield objective.}
The frozen LILAC surrogate predicts the log neutron yield from the generated pulse and laser parameters.
The loss maximizes relative yield gain over the baseline:
\begin{equation}
    \mathcal{L}_{\mathrm{yield}} = -\frac{\exp\!\bigl(S_\phi(x_0, c_{\mathrm{fix}})_0\bigr) - Y_{\mathrm{base}}}{Y_{\mathrm{base}} + \epsilon},
\end{equation}
where $S_\phi(\cdot)_0$ is the predicted log yield (inverse-scaled to physical units) and $Y_{\mathrm{base}} = \exp(S_\phi(x_{\mathrm{ref}}, c_{\mathrm{fix}})_0)$.

\paragraph{Spectral stability.}
The optimized pulse must preserve the frequency content of the reference to ensure physical plausibility:
\begin{equation}
    \mathcal{L}_{\mathrm{stab}} = \frac{1}{N}\sum_{i=1}^{N}\bigl(|\mathcal{F}(x_0)_i| - |\mathcal{F}(x_{\mathrm{ref}})_i|\bigr)^2,
\end{equation}
where $\mathcal{F}$ denotes the discrete Fourier transform over $N$ frequency bins.

\paragraph{Energy conservation.}
The pulse integral must not exceed the reference (one-sided penalty):
\begin{equation}
    \mathcal{L}_{\mathrm{energy}} = \Bigl[\int x_0(t) \cdot s_y\,dt - \int x_{\mathrm{ref}}(t) \cdot s_y\,dt\Bigr]_+,
\end{equation}
where $s_y$ maps normalized values to physical units.

\paragraph{Smoothness.}
Squared first-difference penalty (discrete Dirichlet energy): $\mathcal{L}_{\mathrm{smooth}} = \frac{1}{d-1}\sum_{i}(x_{0,i+1} - x_{0,i})^2$.

\paragraph{Validity criteria.}
The hard constraint thresholds used for candidate selection are detailed in Appendix~\ref{sec:domain_specific_filtering}.

\subsection{Data Center Workload Scheduling}

The inference-time loss for DC combines three terms with Dirichlet-sampled weights $\boldsymbol\lambda = (\lambda_c, \lambda_p, \lambda_{\mathrm{sm}})$:
\begin{equation}
    \mathcal{L}^{\mathrm{DC}} = \lambda_c\,\mathcal{L}_{\mathrm{cost}} + \lambda_p\,\mathcal{L}_{\mathrm{power}} + \lambda_{\mathrm{sm}}\,\mathcal{L}_{\mathrm{smooth}}.
\end{equation}

\paragraph{Cost objective.}
The frozen DC surrogate predicts energy metrics from the generated profile and environmental conditioning.
The loss maximizes relative reduction in the target metric (e.g., carbon footprint $E_9$ or total power $E_0$):
\begin{equation}
    \mathcal{L}_{\mathrm{cost}} = -\frac{f_{\mathrm{base}} - S_\phi(x_0, c_{\mathrm{fix}})_k}{f_{\mathrm{base}} + \epsilon},
\end{equation}
where $S_\phi(\cdot)_k$ is the $k$-th predicted metric inverse-scaled to physical units and $f_{\mathrm{base}}$ is its baseline value.

\paragraph{Workload conservation.}
The total workload delivered must match the reference within a tight tolerance $\tau$:
\begin{equation}
    \mathcal{L}_{\mathrm{power}} = d^2 + \mu\,[\,|d| - \tau\,]_+^2, \qquad d = \frac{\mathcal{I}(x_0) - \mathcal{I}(x_{\mathrm{ref}})}{|\mathcal{I}(x_{\mathrm{ref}})| + \epsilon},
\end{equation}
where $\mathcal{I}(x) = \int x(t) \cdot M\,dt$ is the scaled integral and $\mu$ is the barrier penalty coefficient ($\mu = 100$ in all DC experiments). At inference, in contrast, we optimize against an external carbon objective, so we use a near-equality penalty around the reference workload to prevent the optimizer from exploiting either over-delivery or under-delivery as an artificial route to lower emissions.

\paragraph{Smoothness.}
First- and second-order finite differences:
\begin{equation}
    \mathcal{L}_{\mathrm{smooth}} = \frac{1}{d{-}1}\sum_{i}(\Delta x_i)^2 + \frac{1}{2(d{-}2)}\sum_{i}(\Delta^2 x_i)^2,
\end{equation}
where $\Delta x_i = x_{0,i+1} - x_{0,i}$ and $\Delta^2 x_i = x_{0,i+2} - 2x_{0,i+1} + x_{0,i}$.

\paragraph{Validity criteria.}
The hard constraint thresholds used for candidate selection are detailed in Appendix~\ref{sec:domain_specific_filtering}.

\section{Dirichlet-Weighted Multi-Objective Sampling}\label{app:dirichlet}

The Dirichlet distribution $\mathrm{Dir}(\boldsymbol\alpha)$ is a continuous probability distribution over the $(K{-}1)$-dimensional simplex $\Delta^{K-1} = \{\boldsymbol\lambda \in \mathbb{R}^K : \lambda_k \geq 0,\; \sum_k \lambda_k = 1\}$.
It is parameterized by a concentration vector $\boldsymbol\alpha = (\alpha_1, \ldots, \alpha_K)$ with $\alpha_k > 0$, and has density:
\begin{equation}
    p(\boldsymbol\lambda \mid \boldsymbol\alpha) = \frac{\Gamma\!\bigl(\sum_k \alpha_k\bigr)}{\prod_k \Gamma(\alpha_k)} \prod_{k=1}^K \lambda_k^{\alpha_k - 1},
\end{equation}
where $\Gamma(\cdot)$ is the gamma function.
Key properties relevant to our use:

\begin{itemize}[leftmargin=2em,itemsep=2pt]
    \item \textbf{Simplex constraint.} Every sample $\boldsymbol\lambda \sim \mathrm{Dir}(\boldsymbol\alpha)$ satisfies $\sum_k \lambda_k = 1$ and $\lambda_k \geq 0$ by construction, making it a natural distribution over loss-weight vectors.
    \item \textbf{Symmetric case ($\alpha_k = \alpha$ for all $k$).} When all concentration parameters are equal, the distribution is symmetric over all $K$ objectives. With $\alpha = 1$ (our default), the density is \emph{uniform} over the simplex: every possible trade-off among the $K$ loss terms is equally likely.
    \item \textbf{Concentration control.} Setting $\alpha < 1$ concentrates samples near the vertices of the simplex (i.e., each draw emphasizes a single objective strongly), while $\alpha > 1$ concentrates samples near the center $\boldsymbol\lambda = (1/K, \ldots, 1/K)$ (i.e., more balanced trade-offs). The choice $\alpha = 1$ provides maximum diversity.
\end{itemize}

\paragraph{Intuition for D2C.}
In our framework, each of the $N_\lambda$ weight vectors $\boldsymbol\lambda^{(m)} \sim \mathrm{Dir}(\mathbf{1}_K)$ defines a different scalarization of the multi-objective loss (Eq.~\ref{eq:total_loss}).
Some draws will heavily weight the surrogate yield objective (exploring aggressive performance gains); others will emphasize constraint penalties (producing conservative but highly feasible designs); yet others will balance smoothness against performance.
By optimizing $N_\lambda$ trajectories in parallel with different weight vectors, the candidate pool explores diverse scalarized trade-offs without requiring manual tuning of fixed weights.
This uniformity is with respect to the weight simplex over these task-specific scalarizations; it is not a claim that Dirichlet weighting alone removes arbitrary objective-scale mismatches without the loss design described above.
The validity-first selection (Stage~3 of Algorithm~\ref{alg:d2c}) then picks the best feasible candidate from this diverse pool, combining broad exploration with strict constraint satisfaction.

\section{Theoretical Analysis: Convergence and Stability}

We summarize several local and conditional analytical properties of D2C. The results below establish an exact chain-rule decomposition for gradients through the DDIM reverse chain, a standard gradient-descent bound for a fixed smooth-region subproblem, almost-everywhere differentiability for smooth one-sided penalties, and a simple conditional ranking guarantee for the frozen surrogate objective.

\subsection{Gradient Stability Through the DDIM Chain}

A primary concern in differentiating through iterative denoising chains is the potential for vanishing or exploding gradients as the number of steps $T$ grows. Rather than claiming a full global stability theorem, we first record the exact chain-rule decomposition that underlies gradient flow through the DDIM reverse process.

\textbf{Proposition 1 (Conditioning-Jacobian Decomposition).}
\textit{Let $G_\theta(c, x_T) = x_0$ denote the $T$-step DDIM reverse chain (Eq.~\ref{eq:ddim_step}) with frozen parameters $\theta$, and assume each reverse step is differentiable in $(x_t, c)$. Then:}
\begin{equation}\label{eq:jacobian_decomp}
\frac{\partial x_0}{\partial c} = \sum_{t=1}^{T} \left(\prod_{s=1}^{t-1} J_s \right) \frac{\partial x_{t-1}}{\partial c}\bigg|_{x_t},
\end{equation}
\textit{where $J_s = \frac{\partial x_{s-1}}{\partial x_s}$ is the per-step transition Jacobian, and the direct term $\frac{\partial x_{t-1}}{\partial c}\big|_{x_t}$ captures the local effect of $c$ at step $t$.}

\textbf{Proof.}
Each DDIM step (Eq.~\ref{eq:ddim_step}) computes $x_{t-1} = \phi(x_t, c, t)$ where $\phi$ depends on $c$ through the CFG-guided noise $\tilde\epsilon_\theta$ (Eq.~\ref{eq:cfg}). Applying the chain rule recursively over the $T$ reverse steps yields Eq.~\eqref{eq:jacobian_decomp}. $\square$

\textbf{Remark.}
Proposition~1 is only an exact chain-rule decomposition. A separate heuristic interpretation is that, under a clipped schedule and moderate step sizes, each DDIM transition is close to identity, so recent reverse steps may dominate the gradient more than distant ones. We do not promote that heuristic to a theorem here; we mention it only because it is consistent with the empirical finding in Section~\ref{sec:ablation_inference} that moderate values of $T$ already perform well.

\subsection{Convergence of the Weighted-Sum Optimization}

We now state a standard smooth-nonconvex gradient-descent bound for a single fixed scalarization inside one smooth region of the objective.

\textbf{Proposition 2 (Gradient-Descent Bound Within a Fixed Smooth Region).}
\textit{Let $\mathcal{L}(c) = \sum_{k=1}^K \lambda_k \mathcal{L}_k\bigl(G_\theta(c, x_T)\bigr)$ be the scalarized objective (Eq.~\ref{eq:total_loss}) with fixed weights $\boldsymbol\lambda > 0$ and fixed latent $x_T$. Let $\mathcal{U} \subseteq \mathbb{R}^{d_c}$ be an open region on which $\mathcal{L}$ is $\beta$-smooth and bounded below by $\mathcal{L}^*$. If gradient descent with step size $\eta \leq 1/\beta$ is initialized in $\mathcal{U}$ and all iterates remain in $\mathcal{U}$, then:}
\begin{equation}
\min_{j=0,\ldots,J-1} \|\nabla_c \mathcal{L}(c^{(j)})\|^2 \;\leq\; \frac{2\bigl(\mathcal{L}(c^{(0)}) - \mathcal{L}^*\bigr)}{\eta J}.
\end{equation}
\textit{This is the standard $\mathcal{O}(1/J)$ first-order bound for gradient descent on the restricted smooth-region objective.}

\textbf{Proof.}
This follows from the standard descent lemma for smooth nonconvex optimization~\citep{nesterov2004introductory}. By $\beta$-smoothness: $\mathcal{L}(c^{(j+1)}) \leq \mathcal{L}(c^{(j)}) - \eta(1 - \eta\beta/2)\|\nabla_c \mathcal{L}(c^{(j)})\|^2$. For $\eta \leq 1/\beta$, each step provides descent of at least $(\eta/2)\|\nabla_c \mathcal{L}\|^2$. Telescoping over $J$ steps yields the bound. $\square$

\textbf{Remark.}
Proposition~2 is intentionally conditional and local. It applies only to a single trajectory with fixed $x_T$, fixed weight vector $\boldsymbol\lambda$, plain gradient descent, and iterates that remain inside one pre-specified smooth region $\mathcal{U}$. In the actual implementation, Adam replaces vanilla gradient descent, the optimizable conditioning coordinates are clipped to $[0,1]$ after each step, the loss is only piecewise smooth because of hinge/max/min operators, and different trajectories use independently sampled Dirichlet weights. The proposition therefore should not be read as a global convergence guarantee for the full D2C objective.

Within that idealized smooth-region setting, a stationary point satisfies
\begin{equation}
\sum_{k=1}^K \lambda_k \nabla_c \mathcal{L}_k = 0.
\end{equation}
This is the first-order stationarity condition for the scalarized objective associated with that particular weighting $\boldsymbol\lambda$. Sampling multiple weight vectors $\boldsymbol\lambda$ therefore provides a practical way to generate a diverse set of scalarized trade-offs, but it does not by itself establish broad Pareto-front coverage or any global Pareto-optimality guarantee.

\textbf{Lemma 1 (Smooth One-Sided Penalties Are Differentiable Almost Everywhere).}
\textit{Let $g: \mathbb{R}^d \to \mathbb{R}$ be continuously differentiable, and assume $G_\theta(c, x_T)$ is continuously differentiable in $c$. Define $\mathcal{L}_{\mathrm{fid}}(c) = [g(G_\theta(c, x_T)) - g(x_{\mathrm{ref}})]_+$ and the boundary set $\partial\mathcal{F} = \{c : g(G_\theta(c, x_T)) = g(x_{\mathrm{ref}})\}$. Then $\mathcal{L}_{\mathrm{fid}}$ is differentiable on $\mathbb{R}^{d_c} \setminus \partial\mathcal{F}$, with:}
\begin{equation}
\nabla_c \mathcal{L}_{\mathrm{fid}}(c) = \begin{cases}
\left(\dfrac{\partial x_0}{\partial c}\right)^{\!T} \nabla_{x_0} g & \text{if } g(x_0) > g(x_{\mathrm{ref}}), \\[6pt]
0 & \text{if } g(x_0) < g(x_{\mathrm{ref}}).
\end{cases}
\end{equation}
\textit{Moreover, if $g \circ G_\theta$ is a submersion at every boundary point (i.e., $\nabla_c (g \circ G_\theta)(c) \neq 0$ for all $c \in \partial\mathcal{F}$), then $\partial\mathcal{F}$ is locally a $(d_c{-}1)$-dimensional manifold and in particular has Lebesgue measure zero. Consequently, $\mathcal{L}_{\mathrm{fid}}$ is differentiable almost everywhere.}

\textbf{Proof.}
The function $h(c) = g(G_\theta(c, x_T)) - g(x_{\mathrm{ref}})$ is continuously differentiable since $g$ and $G_\theta$ are. Since $[r]_+ = r \cdot \mathbf{1}_{r > 0}$ is differentiable for $r \neq 0$, the composition $\mathcal{L}_{\mathrm{fid}} = [h(c)]_+$ is differentiable wherever $h(c) \neq 0$, i.e., on $\mathbb{R}^{d_c} \setminus \partial\mathcal{F}$.

In the violated regime ($h(c) > 0$): $\nabla_c \mathcal{L}_{\mathrm{fid}} = \nabla_c h = \left(\frac{\partial x_0}{\partial c}\right)^{\!T} \nabla_{x_0} g$. If additionally $\nabla_c h(c) \neq 0$, then the negative gradient is a nonzero local descent direction for the penalty.

In the feasible interior ($h(c) < 0$): $\mathcal{L}_{\mathrm{fid}} = 0$ identically, so $\nabla_c \mathcal{L}_{\mathrm{fid}} = 0$ and the penalty is inactive.

For the measure-zero claim: the preimage theorem states that if $\nabla_c h(c) \neq 0$ on the level set $h^{-1}(0)$, then $\partial\mathcal{F} = h^{-1}(0)$ is a smooth $(d_c{-}1)$-dimensional submanifold and hence has $d_c$-dimensional Lebesgue measure zero. This establishes almost-everywhere differentiability of $\mathcal{L}_{\mathrm{fid}}$, but it does not imply that optimization trajectories cannot reach $\partial\mathcal{F}$; if an iterate lands on the hinge boundary in practice, first-order optimization is interpreted through the corresponding subgradient. $\square$

\textbf{Remark.}
Lemma~1 applies only to the smooth one-sided penalties obtained from continuously differentiable constraint functionals $g$. Some instantiated fidelity terms in Appendix~\ref{app:training_losses} additionally use coordinatewise $\max/\min$ operators to enforce amplitude bounds; those terms are Lipschitz but not $C^1$, so the manifold and almost-everywhere differentiability claims above should not be read as applying to those non-smooth pieces. In practice, such terms are handled through the corresponding subgradients supplied by automatic differentiation.

\subsection{Limits of the Frozen Surrogate Objective}
The preceding results establish differentiability and local first-order properties of the conditioning optimization, but they do not imply that optimizing the frozen surrogate is equivalent to optimizing the true objective. The relevant question is instead whether the surrogate remains accurate on the region of design space reached through conditioning optimization.

\textbf{Proposition 3 (Pairwise Ordering Reliability Under Uniform Surrogate Error).}
\textit{Fix a latent noise draw $x_T$ and write $x(c) = G_\theta(c, x_T)$. Let $\hat f(x) = S_\phi(x, c_{\mathrm{fix}})$ denote the surrogate objective and let $f(x)$ denote the corresponding true objective. Let $\mathcal{R} \subseteq \{x(c) : c \in \mathbb{R}^{d_c}\}$ be any set that contains the optimization trajectory under consideration. Suppose that}
\begin{equation}
\sup_{x \in \mathcal{R}} |\hat f(x) - f(x)| \leq \varepsilon.
\end{equation}
\textit{Then for any $x_1, x_2 \in \mathcal{R}$,}
\begin{equation}
\big|\big(\hat f(x_1) - \hat f(x_2)\big) - \big(f(x_1) - f(x_2)\big)\big| \leq 2\varepsilon.
\end{equation}
\textit{In particular, if $\hat f(x_1) \geq \hat f(x_2) + 2\varepsilon$, then $f(x_1) \geq f(x_2)$.}

\textbf{Proof.}
By direct expansion,
\begin{equation}
\big(\hat f(x_1) - \hat f(x_2)\big) - \big(f(x_1) - f(x_2)\big)
= \big(\hat f(x_1) - f(x_1)\big) - \big(\hat f(x_2) - f(x_2)\big).
\end{equation}
Applying the triangle inequality and the uniform error bound gives
\begin{equation}
\big|\big(\hat f(x_1) - \hat f(x_2)\big) - \big(f(x_1) - f(x_2)\big)\big|
\leq |\hat f(x_1) - f(x_1)| + |\hat f(x_2) - f(x_2)| \leq 2\varepsilon.
\end{equation}
The ordering statement follows immediately. $\square$

\textbf{Remark.}
This proposition is intentionally conditional. Freezing $\theta$ and $\phi$ does not, by itself, rule out input-space exploitation: optimizing $c$ can still move $x(c)$ into parts of the generator image where the surrogate is inaccurate. What freezing changes is the search space, not the existence of blind spots. D2C instead relies on a weaker and more defensible claim: the frozen diffusion prior, physics-informed penalties, and validity-first filtering can reduce exploitation in practice by biasing search toward designs that remain closer to the empirical data distribution and satisfy explicit feasibility checks. These mechanisms are practical safeguards, not a general theoretical guarantee against surrogate exploitation.

\section{Practical Guidance for Design Optimization with D2C}\label{app:practical_guidance}

D2C is designed for settings where (i) evaluations of the true objective are expensive (physical simulations, wet-lab experiments, or large-scale facility tests), (ii) hard physical constraints must be satisfied for safety or feasibility, and (iii) offline datasets of moderate size are available. Below we summarize the settings where the method is most compelling, together with regimes where simpler alternatives may sometimes be sufficient.

\paragraph{Settings Well Suited to D2C.}
\begin{itemize}[leftmargin=2em,itemsep=2pt]
    \item \textbf{High-cost physical systems with expensive evaluations.} D2C is most valuable when each ground-truth evaluation costs hours of compute (e.g., radiation-hydrodynamics simulations) or involves irreversible physical consequences (e.g., laser shots on a fusion facility, equipment management in data centers). The offline training investment is amortized over many test-time optimizations.
    \item \textbf{Multi-objective problems with hard constraints.} The framework naturally accommodates competing objectives and strict feasibility requirements through the composite loss and validity-first selection. If your application has safety-critical constraints that cannot be violated, D2C provides explicit mechanisms to enforce them.
    \item \textbf{High-dimensional or strongly structured designs.} The diffusion proposer excels at generating designs with complex internal structure (temporal profiles, spatial fields) that naive gradient ascent in design space often cannot capture robustly. If the design space has strong correlations or manifold structure, the learned prior is especially beneficial.
    \item \textbf{Settings where a reliable surrogate exists.} The method requires a differentiable surrogate $S_\phi$ that is accurate in the region of interest. If surrogate error is low ($<5\%$ relative error), the gradient signal is informative.
\end{itemize}

\paragraph{Settings Where Simpler Alternatives May Suffice.}
\begin{itemize}[leftmargin=2em,itemsep=2pt]
    \item \textbf{Low-dimensional problems with weak structure or cheap evaluation.} If the design space is very small, the constraints are mild, and evaluations are cheap enough to support direct search or standard Bayesian optimization, then methods such as GP-qEI, CMA-ES, or even brute-force search may be simpler to deploy and sufficiently effective. This is not a blanket limitation of D2C: even in lower-dimensional settings, the method can still be worthwhile when evaluations are expensive, feasibility is safety-critical, or data-distribution-aware generation and validity-first selection materially improve robustness.
    \item \textbf{Cheap online evaluations available.} If the true objective can be queried cheaply and repeatedly, online methods (reinforcement learning, active learning) are often preferable because they can adapt directly to fresh evaluations. D2C is designed for settings with no online evaluation access at test time.
    \item \textbf{Severely miscalibrated surrogates.} If the surrogate is unreliable in the optimization region, gradient-based methods can exploit those inaccuracies. The diffusion prior and validity checks mitigate this to some extent by constraining designs toward the empirical data distribution, but a fundamentally miscalibrated surrogate can still misdirect optimization.
    \item \textbf{Latency-critical applications.} D2C requires backpropagation through a $T$-step DDIM chain at each of $J$ optimization iterations. On current hardware, this takes 30--60 seconds per test sample. If sub-second inference is required, single-pass methods (MINs, DDOM) may be a better fit despite their weaker optimization and constraint-handling performance.
\end{itemize}

\paragraph{Practical considerations.}
\begin{itemize}[leftmargin=2em,itemsep=2pt]
    \item \textbf{Surrogate investment.} We recommend training the surrogate to convergence with proper validation, achieving $<2\%$ error before proceeding to diffusion model training.
    \item \textbf{Guidance scale sensitivity.} The guidance scale $w$ is domain-dependent and must be tuned. Domains with tight physical constraints (e.g., ICF) are highly sensitive to $w > 2$; smoother landscapes (e.g., DC) are more tolerant. Start with $w = 2$ and sweep $\{1, 2, 3, 5\}$. The sweep does not require oracle access: ranking $w$ by the surrogate-predicted gain of candidates that pass the geometric validity checks recovers the oracle-selected value (Section~\ref{sec:ablation_guidance_selection}).
    \item \textbf{Noise trials vs.\ optimization steps.} When compute budget is fixed, we find it more effective to increase noise trials $N_z$ (broader coverage) than optimization steps $J$ (deeper optimization per trial). Diminishing returns on $J$ set in after 50--100 steps for most problems.
    \item \textbf{Constraint design.} Hard constraints should be expressed as differentiable penalty terms in the loss wherever possible. Non-differentiable constraints (e.g., discrete feasibility checks) should be handled via the post-hoc validity filter rather than incorporated into the gradient-based loop.
\end{itemize}

\section{Broader Impact}
As a general offline design-optimization method, D2C could have positive impact in domains where evaluations are expensive and constrained operation matters, such as reducing carbon-footprint in data centers or accelerating scientific design loops while lowering simulation cost. More broadly, methods that keep optimization closer to the empirical data distribution captured by a learned prior while exposing explicit validity checks help reduce unrealistic proposals relative to unconstrained surrogate optimization.

Potential negative impacts are also application-specific. The same optimization machinery could be used in domains where accelerating design search is undesirable or access should be restricted.

%%%%%%%%%%%%%%%%%%%%%%%%%%%%%%%%%%%%%%%%%%%%%%%%%%%%%%%%%%%%

\newpage
\section*{NeurIPS Paper Checklist}

\begin{enumerate}

\item {\bf Claims}
    \item[] Question: Do the main claims made in the abstract and introduction accurately reflect the paper's contributions and scope?
    \item[] Answer: \answerYes{}
    \item[] Justification: See the abstract, Section~1, and Sections~4--5; the claims are scoped to offline physics-informed optimization and the reported experiments.

\item {\bf Limitations}
    \item[] Question: Does the paper discuss the limitations of the work performed by the authors?
    \item[] Answer: \answerYes{}
    \item[] Justification: Section~5 discusses computational cost, surrogate dependence, validity-first selection, and domain-specific tuning.

\item {\bf Theory assumptions and proofs}
    \item[] Question: For each theoretical result, does the paper provide the full set of assumptions and a complete (and correct) proof?
    \item[] Answer: \answerNA{}
    \item[] Justification: The paper does not include theoretical results.

    \item {\bf Experimental result reproducibility}
    \item[] Question: Does the paper fully disclose all the information needed to reproduce the main experimental results of the paper to the extent that it affects the main claims and/or conclusions of the paper (regardless of whether the code and data are provided or not)?
    \item[] Answer: \answerYes{}
    \item[] Justification: Sections~4--6 and the appendix specify datasets, splits, losses, hyperparameters, baselines, evaluation, and ablation settings.

\item {\bf Open access to data and code}
    \item[] Question: Does the paper provide open access to the data and code, with sufficient instructions to faithfully reproduce the main experimental results, as described in supplemental material?
    \item[] Answer: \answerYes{}
    \item[] Justification: We provide the code used for DC and design-bench attached in the supplementary material. The ICF data and its artifacts are export controlled and cannot be publicly shared (Appendix, ICF data generation).

\item {\bf Experimental setting/details}
    \item[] Question: Does the paper specify all the training and test details (e.g., data splits, hyperparameters, how they were chosen, type of optimizer) necessary to understand the results?
    \item[] Answer: \answerYes{}
    \item[] Justification: Section~4 and Appendix~A report the splits, architectures, optimizers, hyperparameters, and model-selection procedures.

\item {\bf Experiment statistical significance}
    \item[] Question: Does the paper report error bars suitably and correctly defined or other appropriate information about the statistical significance of the experiments?
    \item[] Answer: \answerYes{}
    \item[] Justification: The main results and ablations report mean $\pm$ standard deviation over 5 runs; see Section~4.3 and the table captions.

\item {\bf Experiments compute resources}
    \item[] Question: For each experiment, does the paper provide sufficient information on the computer resources (type of compute workers, memory, time of execution) needed to reproduce the experiments?
    \item[] Answer: \answerYes{}
    \item[] Justification: Appendix~A reports the hardware (four NVIDIA H100 GPUs) and approximate training/optimization costs.
    
\item {\bf Code of ethics}
    \item[] Question: Does the research conducted in the paper conform, in every respect, with the NeurIPS Code of Ethics \url{https://neurips.cc/public/EthicsGuidelines}?
    \item[] Answer: \answerYes{}
    \item[] Justification: To the best of our knowledge, yes; the work uses offline data/simulations, involves no human subjects, and discloses relevant release limitations.

\item {\bf Broader impacts}
    \item[] Question: Does the paper discuss both potential positive societal impacts and negative societal impacts of the work performed?
    \item[] Answer: \answerYes{}
    \item[] Justification: The appendix now includes a short broader-impact discussion covering potential benefits, misuse risks, and deployment risks of the method.
    
\item {\bf Safeguards}
    \item[] Question: Does the paper describe safeguards that have been put in place for responsible release of data or models that have a high risk for misuse (e.g., pre-trained language models, image generators, or scraped datasets)?
    \item[] Answer: \answerNA{}
    \item[] Justification: The submission does not release a high-risk model or scraped dataset.

\item {\bf Licenses for existing assets}
    \item[] Question: Are the creators or original owners of assets (e.g., code, data, models), used in the paper, properly credited and are the license and terms of use explicitly mentioned and properly respected?
    \item[] Answer: \answerYes{}
    \item[] Justification: The paper credits the public code assets used and explicitly states the Design-Bench and SustainDC repository licenses in the appendix data section.

\item {\bf New assets}
    \item[] Question: Are new assets introduced in the paper well documented and is the documentation provided alongside the assets?
    \item[] Answer: \answerNA{}
    \item[] Justification: The paper does not introduce a new publicly released dataset, model, or code asset.

\item {\bf Crowdsourcing and research with human subjects}
    \item[] Question: For crowdsourcing experiments and research with human subjects, does the paper include the full text of instructions given to participants and screenshots, if applicable, as well as details about compensation (if any)? 
    \item[] Answer: \answerNA{}
    \item[] Justification: The paper does not involve crowdsourcing or research with human subjects.

\item {\bf Institutional review board (IRB) approvals or equivalent for research with human subjects}
    \item[] Question: Does the paper describe potential risks incurred by study participants, whether such risks were disclosed to the subjects, and whether Institutional Review Board (IRB) approvals (or an equivalent approval/review based on the requirements of your country or institution) were obtained?
    \item[] Answer: \answerNA{}
    \item[] Justification: The paper does not involve crowdsourcing or research with human subjects.

\item {\bf Declaration of LLM usage}
    \item[] Question: Does the paper describe the usage of LLMs if it is an important, original, or non-standard component of the core methods in this research? Note that if the LLM is used only for writing, editing, or formatting purposes and does \emph{not} impact the core methodology, scientific rigor, or originality of the research, declaration is not required.
    %this research? 
    \item[] Answer: \answerNA{}
    \item[] Justification: The paper does not involve the use of LLMs as an important, original, or non-standard component of the core methods in this research.

\end{enumerate}

\end{document}